\PassOptionsToPackage{table}{xcolor}
\documentclass{article}
\usepackage{iclr2027_conference,times}

\usepackage{amsmath,amsfonts,bm}

\def\eqref#1{equation~\ref{#1}}

\def\1{\bm{1}}

\DeclareMathAlphabet{\mathsfit}{\encodingdefault}{\sfdefault}{m}{sl}
\SetMathAlphabet{\mathsfit}{bold}{\encodingdefault}{\sfdefault}{bx}{n}

\usepackage{amsmath,amssymb}
\usepackage{graphicx}
\usepackage{adjustbox} 
\usepackage{wrapfig}
\usepackage{capt-of}
\usepackage{placeins}
\usepackage{needspace}
\usepackage{booktabs}
\usepackage{tabularx}
\usepackage{array}
\usepackage{longtable}
\usepackage{placeins}
\usepackage{float}
\usepackage[table]{xcolor}

\definecolor{VisionColor}{HTML}{4C78A8}
\definecolor{LanguageColor}{HTML}{F2A541}
\definecolor{ProprioColor}{HTML}{59A14F}
\definecolor{ActionColor}{HTML}{B279A2}
\definecolor{EnvironmentColor}{HTML}{9C9C9C}
\definecolor{DropRed}{HTML}{C44E52}
\definecolor{RobustBlue}{HTML}{4C78A8}
\usepackage{subcaption}
\usepackage{microtype}
\usepackage{hyperref}
\usepackage{url}

\title{LIBERO-MOR: A Multimodal Robustness Benchmark Across the Vision-Language-Action Pipeline}

\author{Qihang Chen$^{1}$, Tengjie Li$^{1}$, Hongwei Lu$^{1}$, Jason Wu$^{2}$, Chyna Qu$^{2}$,
Shikui Tu$^{1*}$, Lei Xu$^{1*}$ \\
$^{1}$School of Computer Science, Shanghai Jiao Tong University \qquad $^{2}$PrismaX Robotics}
\iclrfinalcopy

\newcommand{\figplaceholder}[2]{%
  \fbox{\parbox[c][#2][c]{0.96\linewidth}{\centering\textit{#1}}}%
}

\begin{document}
\maketitle
\lhead{}
\renewcommand{\headrulewidth}{0pt}
{\renewcommand{\thefootnote}{\fnsymbol{footnote}}\footnotetext[1]{Corresponding authors.}}

\begin{abstract}
Vision-language-action (VLA) models have achieved strong performance on robot manipulation benchmarks. Translating this progress into reliable deployment calls for systematic evaluation of how policies respond to variations throughout the robot interaction loop. We introduce \textbf{LIBERO-MOR}, a multimodal robustness benchmark across the vision-language-action pipeline. Built on standard LIBERO, it covers five deployment-facing domains: vision, language, proprioception, action execution, and environment. The benchmark comprises 22 deployment-inspired perturbations, each evaluated at three severity levels across 40 tasks under a paired clean--perturbed protocol, enabling controlled and reproducible comparisons. We evaluate seven representative VLA models spanning diverse action-generation mechanisms. Our experiments reveal substantial robustness gaps behind high clean-task success, together with shared vulnerabilities and distinct model-specific failure patterns across domains. Severity analysis further distinguishes progressive degradation, collapse at stronger interventions, and failures triggered at the mildest tested level. By connecting aggregate performance to pipeline domains, individual perturbations, and intervention strength, LIBERO-MOR provides a systematic framework for diagnosing VLA robustness and guiding the development of more reliable robot policies.
\end{abstract}

\section{Introduction}
\label{sec:intro}
Vision-language-action (VLA) models combine visual and linguistic representations with robot control, using action-generation mechanisms ranging from autoregressive decoding to regression, diffusion, and flow matching \citep{brohan2023rt,kim2024openvla,black2024pi_0,kim2025fine}. Benchmarks such as LIBERO make it possible to compare task completion across policies \citep{liu2023libero}. Deployment-oriented evaluation additionally needs to characterize responses to corrupted observations, incomplete instructions, and delayed commands. These deviations enter different parts of the closed loop, making their location and structure central to understanding policy robustness.

Robustness benchmarks increasingly examine environmental, linguistic, and visual variation, as well as uncertainty across multiple interfaces \citep{pumacay2024colosseum,zhou2025libero,fei2026libero,zhang2025vla,li2026liberovprobenchmarkingclosedloopvisual,guo2026robustness}. Their findings motivate a more specific question: \emph{which intervention sites and perturbation types account for a model's loss of task success, and how does that loss change with intervention strength?} Answering this question requires both broad coverage and a common evaluation protocol. An overall score alone can obscure failures concentrated in one input or control channel.

\begin{figure}[t]
  \centering
  \adjincludegraphics[trim={{.055\width} {.05\height} {.055\width} {.06\height}},clip,width=0.76\textwidth]{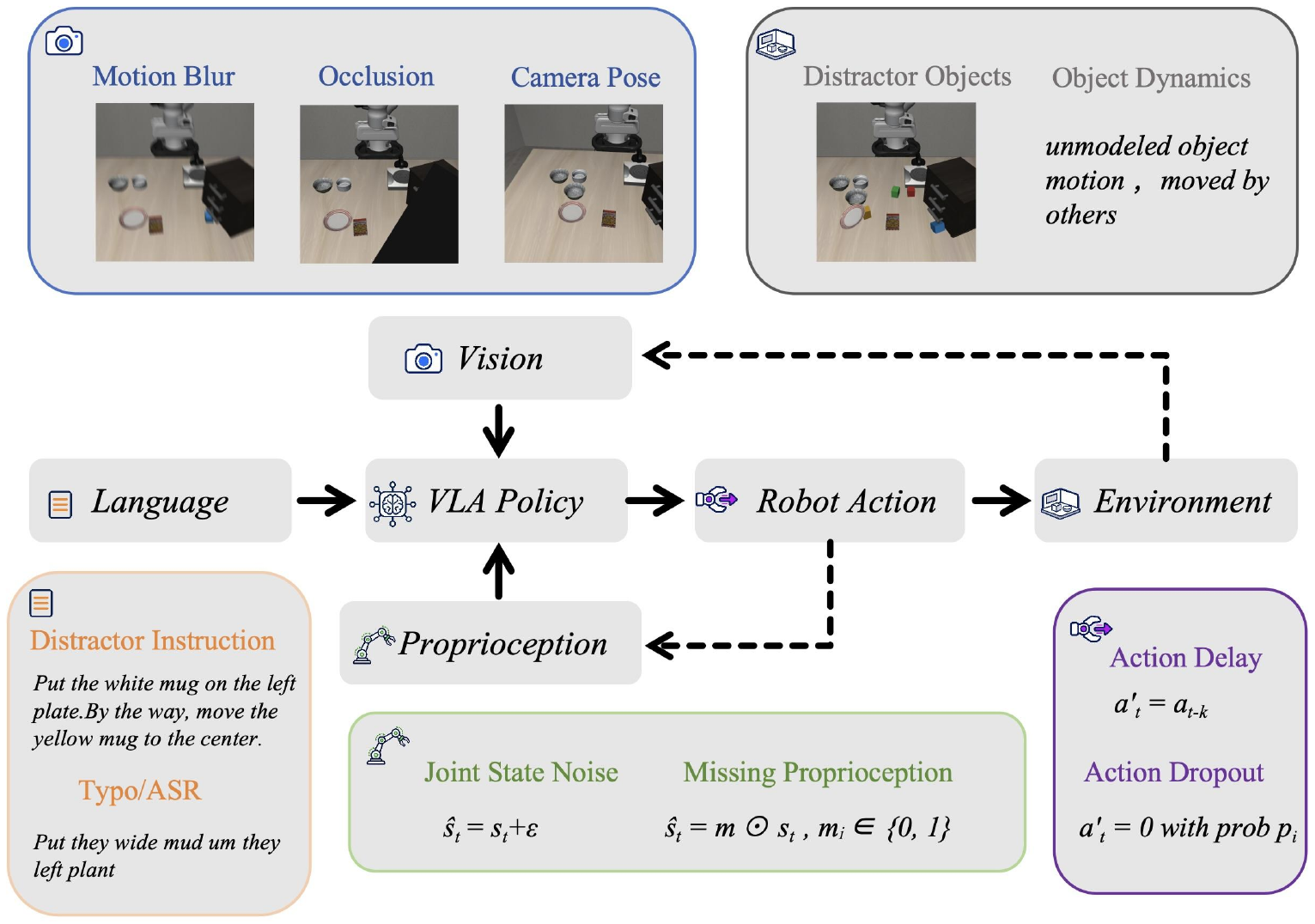}
  \caption{\textbf{LIBERO-MOR overview.} Interventions at five pipeline interfaces preserve the underlying task goal and success checker.}
  \label{fig:overview}
\end{figure}

We introduce \textbf{LIBERO-MOR}, a controlled stress-testing layer on four standard LIBERO suites. Its 22 perturbations cover vision, language, proprioception, action, and environment (Figure~\ref{fig:overview}). A fixed manifest specifies three severity levels and five seeds for each of 40 tasks. Each perturbed episode is linked to a clean task--seed reference, allowing the same protocol to support overall comparisons, domain profiles, perturbation-specific fingerprints, and severity analysis.

We evaluate seven publicly available policies with heterogeneous action-generation designs. Similar clean performance can mask substantial differences under perturbation: GR00T-N1.7 and AVA-VLA both achieve 95.00\% clean success, but their aggregate perturbed success differs by 7.71 percentage points. At a finer resolution, distractor instructions nearly eliminate success for two models, whereas motion blur is especially damaging to another. These patterns motivate a more detailed account of capability than nominal scores alone provide \citep{jiang2026we}.

Our contributions are threefold:
\begin{itemize}\setlength{\itemsep}{1pt}\setlength{\parsep}{0pt}
    \item \textbf{A pipeline-oriented protocol.} LIBERO-MOR evaluates five intervention domains and 22 perturbations with matched clean references, multiple severities, and a fixed task--seed manifest.
    \item \textbf{A comparative evaluation.} We report 93,800 rollouts across seven policies, measuring absolute success, the clean-to-perturbed change, and model-normalized relative robustness.
    \item \textbf{A diagnosis beyond aggregate scores.} Domain profiles, perturbation fingerprints, and severity responses reveal shared vulnerabilities and model-specific failure patterns.
\end{itemize}

\section{Related Work}
\label{sec:related}
\subsection{Vision-Language-Action Models and Evaluation}
VLA designs build on scalable transformer control and vision-language transfer \citep{brohan2022rt,brohan2023rt}, extending to open generalist policies \citep{kim2024openvla,team2024octo}, flow-matching action experts \citep{black2024pi_0,intelligence2025pi_,bjorck2025gr00t}, diffusion action modules \citep{li2024cogact}, latent-action representations \citep{bu2025univla}, efficient continuous action chunking \citep{kim2025fine}, active visual attention \citep{xiao2026ava}, and discrete temporal block diffusion \citep{lee2026tbd}. Our evaluation spans this diversity of policy families and action-generation mechanisms. CALVIN, LIBERO, SIMPLER, RoboCasa, RoboTwin, and VLABench address long-horizon control, transfer, simulated policy evaluation, and task diversity \citep{mees2022calvin,liu2023libero,li2024evaluating,nasiriany2024robocasa,chen2025robotwin,zhang2025vlabench}. LIBERO-MOR complements these capability evaluations by measuring changes relative to each policy's clean performance under the same harness.

\subsection{Robustness Benchmarks for Robot Policies}
The Colosseum, RoboTwin, LIBERO-Pro, LIBERO-Plus, and VLA-Arena study environmental variation, domain randomization, and structured robustness tests \citep{pumacay2024colosseum,chen2025robotwin,zhou2025libero,fei2026libero,zhang2025vla}. LIBERO-VPro focuses on closed-loop visual evidence failures \citep{li2026liberovprobenchmarkingclosedloopvisual}. The closest multimodal precedent is \citet{guo2026robustness}, who evaluate 17 perturbations across action, observation, environment, and instruction uncertainty and introduce a robustness-training method. RoVLA, ProGAL-VLA, and ACoT-VLA further explore training and reasoning strategies for robust control \citep{luo2026rovla,darabi2026progal,zhong2026acot}.

Building on prior multimodal stress tests, LIBERO-MOR explicitly separates proprioception from vision and evaluates 22 perturbations over the same task--seed grid at three intervention strengths. Its contribution lies in the shared diagnostic protocol and cross-model analysis: domain averages identify broad weaknesses, perturbation fingerprints reveal localized failures, and severity curves characterize how those failures develop.

\section{LIBERO-MOR Benchmark}
\label{sec:benchmark}

LIBERO-MOR is a controlled stress-testing layer on standard LIBERO: the underlying task goal and success criterion are preserved while perturbations are injected at specific points of the closed-loop VLA pipeline (Figure~\ref{fig:overview}, Table~\ref{tab:taxonomy}).

\subsection{Benchmark Design Principles}
\label{sec:design-principles}

\textbf{Deployment relevance.} Perturbations represent deployment-facing failures in sensing, instruction, state estimation, control, and the physical scene. Their controlled formulation makes these factors reproducible and supports targeted analysis of policy responses.

\textbf{Pipeline coverage.} We organize failures by where they enter the closed loop: \emph{vision}, \emph{language}, \emph{proprioception}, \emph{action}, and \emph{environment}. The first three modify policy inputs, action perturbations modify commands before execution, and environment perturbations modify the scene or dynamics.

\textbf{Controlled single-factor interventions.} Each perturbed rollout activates one perturbation at one severity while retaining the matched task, seed, goal, and original success checker. This design associates the measured response with a specific injected factor and captures its downstream consequences throughout the closed loop, providing a clear basis for comparing failure sources.

\textbf{Graded severity.} Each perturbation has three ordered intervention strengths $s\in\{1,2,3\}$. We measure the resulting success rates at every level, using the curves to characterize how closed-loop behavior changes with intervention strength.

\subsection{Multimodal Perturbation Taxonomy}
\label{sec:taxonomy}

\begin{table*}[t]
\caption{\textbf{LIBERO-MOR perturbation taxonomy.} Twenty-two perturbations are grouped by their injection site. The five domains include both policy inputs and downstream execution conditions.}
\label{tab:taxonomy}
\centering
\small
\setlength{\tabcolsep}{5.2pt}
\renewcommand{\arraystretch}{1.12}
\begin{tabularx}{\textwidth}{@{}>{\raggedright\arraybackslash}p{1.78cm} >{\centering\arraybackslash}p{0.38cm} X X@{}}
\toprule
\textbf{Domain} & \textbf{\#} & \textbf{Perturbations} & \textbf{Deployment factors represented} \\
\midrule
\rowcolor{VisionColor!10}
\cellcolor{VisionColor!24}\textbf{Vision} & \textbf{10} &
Gaussian noise; motion blur; defocus blur; compression; brightness; shadow; splash; occlusion; camera shake; camera pose shift &
Sensor degradation, blur, bandwidth loss, illumination change, partial visibility, camera motion, and viewpoint/calibration shift \\
\addlinespace[1.5pt]
\rowcolor{LanguageColor!10}
\cellcolor{LanguageColor!24}\textbf{Language} & \textbf{4} &
Paraphrase; ambiguous instruction; distractor instruction; typo / ASR-like corruption &
Alternative wording, under-specification, competing linguistic content, and transcription-like errors \\
\addlinespace[1.5pt]
\rowcolor{ProprioColor!10}
\cellcolor{ProprioColor!24}\textbf{Proprio.} & \textbf{3} &
Joint-state noise; end-effector pose noise; missing proprioception &
Encoder/state-estimation noise, calibration error, and state-channel loss \\
\addlinespace[1.5pt]
\rowcolor{ActionColor!10}
\cellcolor{ActionColor!24}\textbf{Action} & \textbf{3} &
Action-execution noise; action delay; action dropout &
Low-level execution error, control/communication latency, and dropped commands \\
\addlinespace[1.5pt]
\rowcolor{EnvironmentColor!10}
\cellcolor{EnvironmentColor!24}\textbf{Environment} & \textbf{2} &
Object dynamics; distractor object &
Changes in object motion and task-irrelevant physical objects \\
\bottomrule
\end{tabularx}
\end{table*}

The taxonomy is deliberately interface-centric: observation-side perturbations act before inference, action perturbations act between prediction and execution, and environment perturbations modify simulator state or scene. Domain labels follow the intervention site, which gives a consistent taxonomy even when effects propagate across interfaces. For example, an environment intervention changes the simulated scene and can subsequently alter visual observations. A vision intervention instead operates on the observation channel. Language under-specification changes the information presented to the policy while retaining the original benchmark goal and success checker. This separation between the task objective and the evidence available to the policy enables targeted evaluation of instruction robustness.

\subsection{Evaluation Protocol and Metrics}
\label{sec:protocol}

We instantiate LIBERO-MOR over the 40 tasks in LIBERO-Spatial, LIBERO-Object, LIBERO-Goal, and LIBERO-Long (10 tasks per suite; LIBERO-Long denotes LIBERO-10). Let $\mathcal{T}$ denote the 40 tasks, $\mathcal{P}$ the 22 perturbations, $\mathcal{S}=\{1,2,3\}$ the three severity levels, and $\mathcal{R}$ the five evaluation seeds. The perturbed benchmark therefore contains $|\mathcal{T}|\,|\mathcal{P}|\,|\mathcal{S}|\,|\mathcal{R}|=40\times22\times3\times5=13{,}200$ rollouts per model. Each perturbation contributes 600 episodes (200 per severity); adding one clean rollout per task--seed pair yields 13,400 episodes per model and 93,800 across seven VLAs. The 200 clean episodes are reused as references for all 66 perturbation--severity settings.

For model $m$, we report the clean success rate $\mathrm{SR}^{m}_{\mathrm{clean}}$ over the 200 clean episodes and the aggregate perturbed success rate $\mathrm{SR}^{m}_{\mathrm{pert}}$ over all 13,200 perturbed episodes. Because every perturbation contributes the same number of episodes, the aggregate rate equally weights the 22 perturbation types, the three severities, and the 40 tasks. Each domain contributes in proportion to its number of perturbation types: vision contains 10 types, whereas environment contains two. Domain-level rates average the perturbations within that domain. We summarize robustness degradation using
\begin{equation}
\Delta \mathrm{SR}^{m}
=\mathrm{SR}^{m}_{\mathrm{pert}}-\mathrm{SR}^{m}_{\mathrm{clean}},
\qquad
\mathrm{RR}^{m}
=\frac{\mathrm{SR}^{m}_{\mathrm{pert}}}{\mathrm{SR}^{m}_{\mathrm{clean}}},
\label{eq:robustness-metrics}
\end{equation}
where $\Delta\mathrm{SR}$ is reported in percentage points and RR is the ratio of two aggregate success rates. RR is defined when $\mathrm{SR}^{m}_{\mathrm{clean}}>0$ and is reported as a percentage. RR summarizes performance over the full task--seed grid, including both success-to-failure and failure-to-success transitions relative to clean evaluation. Reporting it alongside both absolute rates separates the achieved level of task success from the proportional change under perturbation.

Domain-, perturbation-, and severity-level RR use the same model-specific clean denominator. Cross-model summaries give equal weight to the seven model-specific ratios. Min--max intervals in Figure~\ref{fig:perturbation-ranking} quantify the spread across evaluated policies.

\section{Evaluation Setup}
\label{sec:setup}

\subsection{Evaluated VLA Models}
\label{sec:models}

We evaluate seven public LIBERO-capable policies---OpenVLA-OFT, $\pi_{0.5}$, UniVLA, GR00T-N1.7-LIBERO, AVA-VLA, TBD-VLA, and CogACT-Libero \citep{kim2025fine,intelligence2025pi_,bu2025univla,nvidia2026groot17,xiao2026ava,lee2026tbd,li2024cogact}---chosen to span heterogeneous action-generation designs (Table~\ref{tab:models}).

\begin{table*}[t]
\caption{\textbf{Evaluated VLA models.} Seven policy families cover diverse representations and action-generation mechanisms, enabling comparisons of robustness profiles across contemporary VLA systems.}
\label{tab:models}
\centering
\small
\setlength{\tabcolsep}{5.0pt}
\renewcommand{\arraystretch}{1.1}
\begin{tabularx}{\textwidth}{@{}>{\raggedright\arraybackslash}p{2.25cm} X X@{}}
\toprule
\textbf{Model} & \textbf{Representative design} & \textbf{Action generation} \\
\midrule
OpenVLA-OFT & Optimized fine-tuning of OpenVLA for efficient continuous robot control & Parallel continuous action-chunk regression \\
\rowcolor{black!3}
$\pi_{0.5}$ & Pretrained vision-language backbone coupled to a specialized action expert & Flow-matched continuous action chunks \\
UniVLA & Cross-embodiment policy built around task-centric latent actions & Latent-action decoding to robot commands \\
\rowcolor{black!3}
GR00T-N1.7-LIBERO & Generalist robot policy with embodiment-conditioned state/action modules & Flow-matching transformer action head \\
AVA-VLA & Task-history modeling with active visual attention & Continuous action prediction \\
\rowcolor{black!3}
TBD-VLA & Discrete temporal block diffusion policy & Masked token diffusion within blocks; autoregression across blocks \\
CogACT-Libero & VLM-based cognition module paired with a specialized action module & Diffusion-transformer action-sequence generation \\
\bottomrule
\end{tabularx}
\end{table*}

We evaluate released or LIBERO-adapted checkpoints with fixed inference configurations. LIBERO-MOR perturbations are introduced exclusively during evaluation. Clean success measured by the same harness provides the reference in Eq.~\ref{eq:robustness-metrics}, aligning nominal and perturbed scores within a common execution protocol.

\subsection{Implementation and Evaluation Protocol}
\label{sec:implementation}

All models share the standard LIBERO simulator \citep{liu2023libero}, fixed manifest, 220-step episode cap, and action horizon of 10 environment steps.
Model-specific adapters map the common interface to native preprocessing, state/action normalization, and output formats while preserving the shared tasks, perturbation operators, dynamics, and success criteria. Manifest audit confirms 200 clean and 13,200 perturbed episodes per model, 200 unique clean keys, 13,200 unique variant IDs, and zero invalid episodes. Execution validity verifies completion of the evaluation protocol, and the original LIBERO checker separately determines task success.

This setup evaluates complete checkpoint--adapter systems under a common stress test. Native preprocessing and normalization connect simulator-side interventions to each policy's representation and action interface. The resulting system-level profiles identify targets for follow-up ablations: concentrated proprioceptive weaknesses motivate testing state inputs, while language-specific failures motivate examining instruction processing under matched task conditions.

\section{Benchmark Results and Analysis}
\label{sec:results}

\subsection{Robustness Degrades Substantially Beyond Clean Evaluation}
\label{sec:overall}

\begin{figure}[t]
\centering
\begin{minipage}[t]{0.505\textwidth}
\vspace{0pt}
\captionof{table}{\textbf{Overall LIBERO-MOR robustness.} Success rates and RR are in \%; $\Delta$SR is in percentage points. Color intensity encodes larger drop (red) or retention (blue).}
\label{tab:overall}
\centering
\small
\setlength{\tabcolsep}{4pt}
\renewcommand{\arraystretch}{1.18}
\begin{tabular*}{\linewidth}{@{\extracolsep{\fill}}lcccc@{}}
\toprule
\textbf{Model} & \textbf{Clean} & \textbf{Pert.} & \textbf{$\Delta$SR} & \textbf{RR} \\
\midrule
OpenVLA-OFT & 92.50 & 48.64 & \cellcolor{DropRed!18}$-43.86$ & \cellcolor{RobustBlue!7}52.58 \\
$\pi_{0.5}$ & \textbf{97.00} & \textbf{62.68} & \cellcolor{DropRed!8}\textbf{$-34.32$} & \cellcolor{RobustBlue!22}\textbf{64.62} \\
UniVLA & 79.00 & 39.68 & \cellcolor{DropRed!14}$-39.32$ & \cellcolor{RobustBlue!5}50.23 \\
GR00T-N1.7 & 95.00 & 57.38 & \cellcolor{DropRed!12}$-37.62$ & \cellcolor{RobustBlue!17}60.40 \\
AVA-VLA & 95.00 & 49.67 & \cellcolor{DropRed!20}$-45.33$ & \cellcolor{RobustBlue!7}52.29 \\
TBD-VLA & 95.50 & 59.50 & \cellcolor{DropRed!10}$-36.00$ & \cellcolor{RobustBlue!20}62.30 \\
CogACT & 94.00 & 58.33 & \cellcolor{DropRed!10}$-35.67$ & \cellcolor{RobustBlue!20}62.06 \\
\bottomrule
\end{tabular*}
\end{minipage}\hfill
\begin{minipage}[t]{0.455\textwidth}
\vspace{0pt}
\centering
\includegraphics[width=\linewidth]{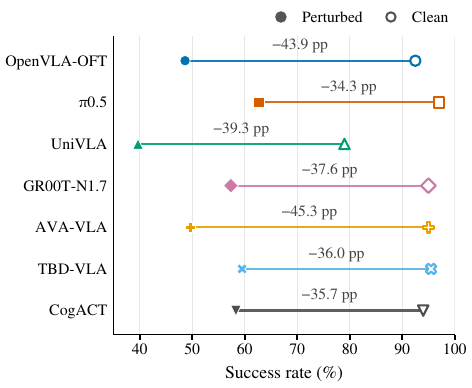}
\caption{\textbf{Clean-to-perturbed degradation.}
Hollow and filled markers denote clean and perturbed SR;
labels show $\Delta$SR (pp).}
\label{fig:overall-drop}
\end{minipage}
\end{figure}

Table~\ref{tab:overall} and Figure~\ref{fig:overall-drop} show substantial degradation for every VLA: clean SR is 79.00--97.00\%, perturbed SR falls to 39.68--62.68\%, and $\Delta$SR spans $-45.33$ to $-34.32$ points. RR ranges from 50.23\% to 64.62\%, quantifying a substantial robustness gap across the evaluated policies.

Perturbation evaluation reveals a different model ordering from clean evaluation. CogACT has slightly lower clean success than AVA-VLA (94.00\% versus 95.00\%) but higher perturbed success (58.33\% versus 49.67\%). Their RR values, 62.06\% and 52.29\%, show that this difference persists after normalization. Conversely, UniVLA's lowest perturbed success coincides with its lowest clean success, illustrating why absolute performance and the relative change should be read together.

\subsection{Which Deployment Perturbations Are Most Challenging?}
\label{sec:global-perturbations}

\begin{figure*}[t]
  \centering
  \IfFileExists{figures/fig3_perturbation_vulnerability_ranking.pdf}{%
    \includegraphics[width=0.96\textwidth]{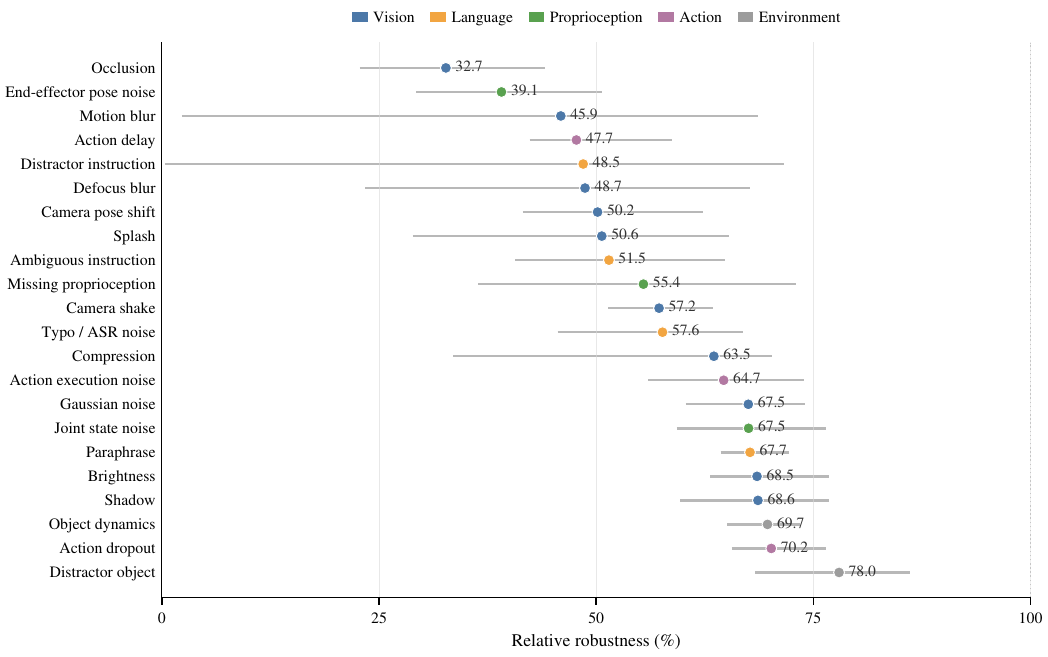}%
  }{\figplaceholder{Figure 3 placeholder: global perturbation vulnerability ranking}{6.2cm}}
  \caption{\textbf{Global perturbation vulnerability.} Dots show mean RR across seven VLAs; intervals denote model-wise minima and maxima, and colors indicate deployment domains.}
  \label{fig:perturbation-ranking}
\end{figure*}

Figure~\ref{fig:perturbation-ranking} ranks individual perturbations by mean model-normalized RR across seven VLAs; the min--max interval exposes cross-model dependence.

Several perturbations reveal \textbf{shared vulnerabilities}. Occlusion is the strongest aggregate failure, with 32.70\% mean RR and a 22.78--44.16\% range across models. End-effector pose noise follows at 39.10\% (29.30--50.69\%), showing that corrupted geometric state can be damaging without directly perturbing vision or language. Action delay is similarly broad: its mean RR is 47.72\%, with all models lying between 42.34\% and 58.76\% across the tested severity mixture.

Other perturbations expose \textbf{strong cross-model disagreement}. Motion blur averages 45.93\% RR but spans 2.32--68.59\%; distractor instruction averages 48.50\% but ranges from 0.36\% to 71.65\%; and defocus blur ranges from 23.42\% to 67.71\%. Thus, similar mean difficulty can conceal model-specific collapse. At the higher-RR end, distractor object reaches 77.96\% mean RR, followed by action dropout (70.16\%) and object dynamics (69.72\%).

\Needspace{5\baselineskip}
\subsection{Robustness Profiles Differ Across Pipeline Domains}
\label{sec:domain-profiles}

\begin{wrapfigure}{L}{0.48\textwidth}
  \centering
  \IfFileExists{figures/fig4_modality_robustness_profiles.pdf}{%
    \includegraphics[width=\linewidth]{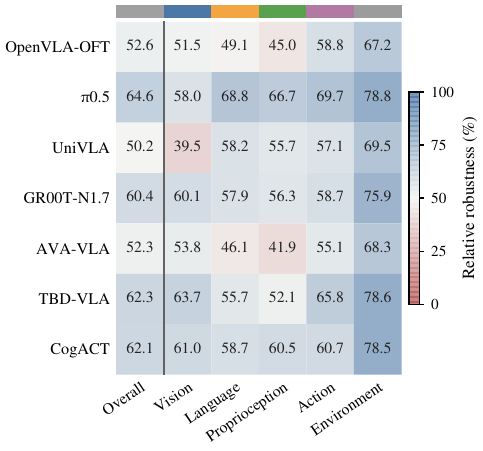}%
  }{\figplaceholder{Figure 4 placeholder: multimodal robustness profiles}{4.0cm}}
  \caption{\textbf{Pipeline robustness profiles.} Cells report model-normalized RR (\%) overall and across five deployment domains.}
  \label{fig:domain-profiles}
\end{wrapfigure}

Aggregating perturbations by pipeline location reveals a complementary axis of robustness variation. Figure~\ref{fig:domain-profiles} reports model-normalized RR for each deployment domain, with overall RR as reference.

Environment has the highest RR for every model (67.2--78.8\% RR). Across models, environment and action average 73.8\% and 60.8\%, whereas vision (55.4\%), language (56.4\%), and proprioception (54.0\%) are more fragile; the domain profiles summarize these family-specific interventions and highlight the pipeline locations associated with the largest performance changes.

Profile shape differs across models. UniVLA is weakest in vision (39.5\% RR), whereas AVA-VLA is weakest in proprioception (41.9\%) and language (46.1\%). $\pi_{0.5}$ retains 66.7--69.7\% across language, proprioception, and action, while TBD-VLA is stronger in vision/action than language/proprioception. GR00T-N1.7 and CogACT are comparatively flat outside environment. Thus, similar aggregate RR can arise from different interface-specific strengths and weaknesses.

\WFclear
\subsection{Fine-Grained Robustness Fingerprints Reveal Distinct Failure Modes}
\label{sec:fingerprints}

\begin{figure*}[t]
  \centering
  \IfFileExists{figures/fig5_perturbation_robustness_fingerprint.pdf}{%
    \includegraphics[width=0.98\textwidth]{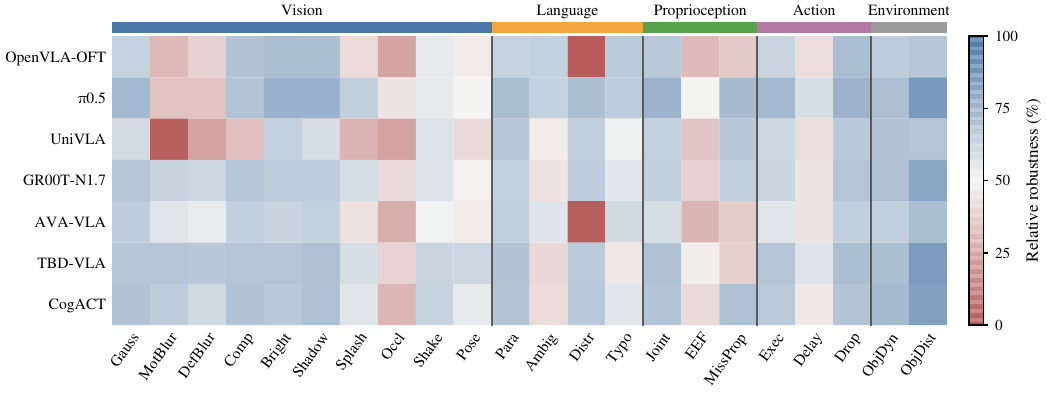}%
  }{\figplaceholder{Figure 5 placeholder: fine-grained perturbation robustness fingerprint}{5.8cm}}
  \caption{\textbf{Fine-grained robustness fingerprints.} Cell colors encode model-normalized RR (\%) for all 22 perturbations, grouped by deployment domain.}
  \label{fig:fingerprint}
\end{figure*}

Domain averages can still mask sharply different failures within the same interface. Figure~\ref{fig:fingerprint} therefore represents each model by a 22-dimensional perturbation-level RR fingerprint.

The clearest example occurs in the language domain. OpenVLA-OFT retains 64.3--66.8\% RR under paraphrase, ambiguous instruction, and typo corruption, yet collapses to 0.4\% under distractor instruction. AVA-VLA exhibits the same qualitative pattern: 56.7--65.3\% RR for the other language perturbations but only 1.4\% for distractor instruction. By contrast, several other models remain above 60\% RR under the same distractor perturbation. The contrast identifies distractor-instruction processing as a specific diagnostic target for examining instruction formatting, attention, and training coverage.

Fine-grained visual and proprioceptive responses are similarly non-uniform. For $\pi_{0.5}$, brightness and shadow each retain 76.8\% RR, whereas motion and defocus blur fall to 34.4\% and 34.2\%. UniVLA shows an even sharper visual fingerprint: Gaussian noise, brightness, and camera shake retain 57.6--64.8\% RR, while motion blur falls to 2.3\% and defocus blur to 23.4\%. Within proprioception, CogACT is comparatively tolerant to joint-state noise and missing proprioception (69.1\% and 70.9\%) but substantially weaker to end-effector pose noise (41.3\%); TBD-VLA instead retains 70.7\% under joint-state noise but only 37.7\% when proprioception is missing.

Under the tested settings, action dropout is less damaging than action delay for every model despite sharing the control channel. The fingerprints separate these temporal disturbance patterns and support selecting mitigation targets at the level of individual control operators.

\subsection{Severity Reveals Qualitatively Different Failure Dynamics}
\label{sec:severity}

\begin{figure*}[t]
  \centering
  \IfFileExists{figures/fig6_severity_failure_dynamics.pdf}{%
    \includegraphics[width=0.94\textwidth]{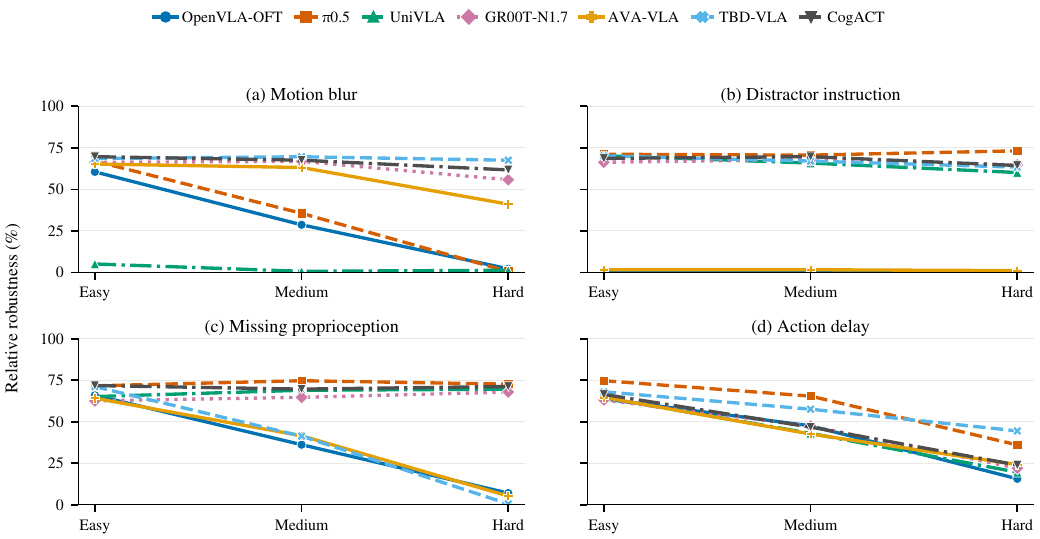}%
  }{\figplaceholder{Figure 6 placeholder: severity-dependent failure dynamics}{7.0cm}}
  \caption{\textbf{Severity-dependent robustness.} Curves show RR at three intervention levels (Easy, Medium, and Hard denote mild, medium, and strong settings). Lines connect the measured response at each intervention level.}
  \label{fig:severity}
\end{figure*}

Severity analysis examines how vulnerabilities change as the intervention setting increases. Figure~\ref{fig:severity} traces model-normalized RR across three levels for four representative perturbations chosen to expose distinct response shapes.

\textbf{Progressive collapse versus a comparatively flat response.}
Motion blur produces sharply different trajectories. OpenVLA-OFT falls from 60.5\% RR at mild severity to 28.6\% at medium and 2.2\% at strong severity; $\pi_{0.5}$ follows a similar 67.0\% $\rightarrow$ 35.6\% $\rightarrow$ 0.5\% trajectory. TBD-VLA instead remains near 68\% across the three levels, while CogACT and GR00T-N1.7 decline less steeply. The approximately flat TBD-VLA curve separates the initial clean-to-perturbed loss from the subsequent response to stronger blur.

Missing proprioception yields a related split. OpenVLA-OFT, AVA-VLA, and TBD-VLA reach 7.0\%, 5.3\%, and 0.5\% RR at the strongest level, whereas $\pi_{0.5}$, UniVLA, GR00T-N1.7, and CogACT remain approximately stable or show small increases. The strongest setting is thus a particularly informative stress test for the first three policies.

\textbf{Failure at the mildest setting versus graded degradation.}
OpenVLA-OFT and AVA-VLA already have only 0.5\% and 1.6\% RR under the mildest distractor instruction and remain near zero at higher levels. The other five models stay around 60--73\%. Thus, the tested distractor format is sufficient to trigger failure in the first two systems without requiring the strongest setting. In contrast, all seven models lose RR as action delay increases, reaching 15.7--44.5\% at strong severity. This shared downward trend identifies delayed execution as a broadly challenging control condition, with model-dependent rates of degradation.

\textbf{Interpreting non-monotonic responses.}
Each severity point summarizes 200 episodes over the same 40 tasks and five seeds. At this sample size, one changed outcome corresponds to 0.5 percentage points of SR before clean-rate normalization. Severity indexes the intervention setting, while empirical success captures the policy's resulting closed-loop response. Small reversals can arise from finite-sample variation, task-specific interactions, or incidental benefits along a trajectory; stronger interventions can also change the distribution of visited states. Paired episode traces and operator-level checks provide a direct route to distinguishing these candidate explanations and locating the tasks or execution stages responsible for a reversal.

\section{Discussion and Conclusion}
\label{sec:discussion}

\subsection{Implications for Robustness Evaluation}
The results support a sequence of reporting decisions. First, report clean and perturbed success under the same execution protocol, so that the absolute loss remains visible. Second, provide RR as a complementary normalization, while retaining its clean denominator. Third, inspect domain and perturbation profiles before selecting a mitigation target. Finally, examine severity curves to distinguish failures already present at the mildest setting from those that worsen across the tested range. The same hierarchy supports evaluating robustness interventions: an improvement can be traced to its affected perturbations and severity levels, while the remaining domains reveal whether the benefit transfers across the pipeline.

The choice of aggregation also matters for deployment-oriented interpretation. LIBERO-MOR uses uniform perturbation-type weights to establish a transparent and reproducible comparison distribution. A deployment with frequent communication delay may prioritize a different profile from one dominated by poor visual conditions. Accordingly, the overall benchmark score summarizes a specified test mixture, while the disaggregated results allow readers to examine the conditions relevant to their application. Aligning operators, severity ranges, checkpoints, and rollout budgets provides a common basis for extending these comparisons to other robustness benchmarks.

\subsection{Scope and Future Directions}
LIBERO-MOR establishes a controlled evaluation setting within the LIBERO embodiment, using family-specific severity scales, single-factor interventions, binary task success, and five seeds per task. The shared execution budget supports systematic comparisons and several directions for extending the benchmark. Recovery time, trajectory quality, and partial progress can enrich the behavioral characterization; expanded seed coverage and episode-level uncertainty estimates can sharpen task-specific comparisons. Controlled ablations of architecture, pretraining, fine-tuning, and adapters can further explain the observed system-level differences. Additional embodiments, compound interventions, policy-native execution settings, and real-robot experiments can test how these robustness profiles transfer across deployment conditions.

\paragraph{Conclusion.}
LIBERO-MOR evaluates perturbations across the VLA control loop with a matched, multi-severity protocol. Across seven policies, substantial aggregate losses coexist with distinct perturbation-specific and severity-dependent responses. These results make the case for reporting \emph{which conditions cause failure and how the response changes with severity}, alongside clean task success.

\clearpage
\subsection*{AI use statement}
The research idea, methodological direction, and overall framework were conceived and determined by the authors. Generative AI tools were used to assist with searching for and summarizing relevant literature based on the authors' proposed research direction, to implement substantial portions of the code after the overall framework had been established by the authors, and to assist with manuscript drafting, language editing, and LaTeX formatting. All AI-assisted outputs, including literature references, code, experimental results, and manuscript content, were reviewed and verified by the authors, who take full responsibility for the final work.

\subsection*{Reproducibility statement}
Section~\ref{sec:benchmark} defines the benchmark composition, including the task--perturbation--severity--seed evaluation grid, matched clean references, and aggregation procedures. Section~\ref{sec:setup} reports the evaluated policy families, shared simulator and execution budget, the scope of model-specific adapters, and manifest audit statistics. Tables~\ref{tab:taxonomy} and~\ref{tab:models} summarize the perturbation taxonomy and the representative policy designs. The code, benchmark implementation, evaluation configurations, and associated datasets and artifacts developed for this project will be released publicly to support reproducibility and future research.

\bibliography{references}

\begin{thebibliography}{29}
\providecommand{\natexlab}[1]{#1}
\providecommand{\url}[1]{\texttt{#1}}
\expandafter\ifx\csname urlstyle\endcsname\relax
  \providecommand{\doi}[1]{doi: #1}\else
  \providecommand{\doi}{doi: \begingroup \urlstyle{rm}\Url}\fi

\bibitem[Bjorck et~al.(2025)Bjorck, Casta{\~n}eda, Cherniadev, Da, Ding, Fan,
  Fang, Fox, Hu, Huang, et~al.]{bjorck2025gr00t}
Johan Bjorck, Fernando Casta{\~n}eda, Nikita Cherniadev, Xingye Da, Runyu Ding,
  Linxi Fan, Yu~Fang, Dieter Fox, Fengyuan Hu, Spencer Huang, et~al.
\newblock Gr00t n1: An open foundation model for generalist humanoid robots.
\newblock \emph{arXiv preprint arXiv:2503.14734}, 2025.

\bibitem[Black et~al.(2024)Black, Brown, Driess, Esmail, Equi, Finn, Fusai,
  Groom, Hausman, Ichter, et~al.]{black2024pi_0}
Kevin Black, Noah Brown, Danny Driess, Adnan Esmail, Michael Equi, Chelsea
  Finn, Niccolo Fusai, Lachy Groom, Karol Hausman, Brian Ichter, et~al.
\newblock {$\pi_0$}: A vision-language-action flow model for general robot
  control.
\newblock \emph{arXiv preprint arXiv:2410.24164}, 2024.

\bibitem[Black et~al.(2025)Black, Brown, Darpinian, Dhabalia, Driess, Esmail,
  Equi, Finn, Fusai, et~al.]{intelligence2025pi_}
Kevin Black, Noah Brown, James Darpinian, Karan Dhabalia, Danny Driess, Adnan
  Esmail, Michael Equi, Chelsea Finn, Niccolo Fusai, et~al.
\newblock {$\pi_{0.5}$}: A vision-language-action model with open-world
  generalization.
\newblock \emph{arXiv preprint arXiv:2504.16054}, 2025.

\bibitem[Brohan et~al.(2022)Brohan, Brown, Carbajal, Chebotar, Dabis, Finn,
  Gopalakrishnan, Hausman, Herzog, Hsu, et~al.]{brohan2022rt}
Anthony Brohan, Noah Brown, Justice Carbajal, Yevgen Chebotar, Joseph Dabis,
  Chelsea Finn, Keerthana Gopalakrishnan, Karol Hausman, Alex Herzog, Jasmine
  Hsu, et~al.
\newblock Rt-1: Robotics transformer for real-world control at scale.
\newblock \emph{arXiv preprint arXiv:2212.06817}, 2022.

\bibitem[Brohan et~al.(2023)Brohan, Brown, Carbajal, Chebotar, Chen,
  Choromanski, Ding, Driess, Dubey, Finn, et~al.]{brohan2023rt}
Anthony Brohan, Noah Brown, Justice Carbajal, Yevgen Chebotar, Xi~Chen,
  Krzysztof Choromanski, Tianli Ding, Danny Driess, Avinava Dubey, Chelsea
  Finn, et~al.
\newblock Rt-2: Vision-language-action models transfer web knowledge to robotic
  control.
\newblock \emph{arXiv preprint arXiv:2307.15818}, 2023.

\bibitem[Bu et~al.(2025)Bu, Yang, Cai, Gao, Ren, Yao, Luo, and
  Li]{bu2025univla}
Qingwen Bu, Yanting Yang, Jisong Cai, Shenyuan Gao, Guanghui Ren, Maoqing Yao,
  Ping Luo, and Hongyang Li.
\newblock Univla: Learning to act anywhere with task-centric latent actions.
\newblock \emph{arXiv preprint arXiv:2505.06111}, 2025.

\bibitem[Chen et~al.(2025)Chen, Chen, Chen, Cai, Liu, Li, Liang, Lin, Ge, Gu,
  et~al.]{chen2025robotwin}
Tianxing Chen, Zanxin Chen, Baijun Chen, Zijian Cai, Yibin Liu, Zixuan Li,
  Qiwei Liang, Xianliang Lin, Yiheng Ge, Zhenyu Gu, et~al.
\newblock Robotwin 2.0: A scalable data generator and benchmark with strong
  domain randomization for robust bimanual robotic manipulation.
\newblock \emph{arXiv preprint arXiv:2506.18088}, 2025.

\bibitem[Darabi \& Trivedi(2026)Darabi and Trivedi]{darabi2026progal}
Nastaran Darabi and Amit~Ranjan Trivedi.
\newblock Progal-vla: Grounded alignment through prospective reasoning in
  vision-language-action models.
\newblock \emph{arXiv preprint arXiv:2604.09824}, 2026.

\bibitem[Fei et~al.(2026)Fei, Wang, Shi, Dai, Cai, Qian, Ji, He, Zhang, Fei,
  et~al.]{fei2026libero}
Senyu Fei, Siyin Wang, Junhao Shi, Zihao Dai, Jikun Cai, Pengfang Qian, Li~Ji,
  Xinzhe He, Shiduo Zhang, Zhaoye Fei, et~al.
\newblock Libero-plus: A progressive robustness benchmark for
  visual-language-action models.
\newblock In \emph{Proceedings of the IEEE/CVF Conference on Computer Vision
  and Pattern Recognition}, pp.\  38574--38583, 2026.

\bibitem[Guo et~al.(2026)Guo, Wu, Tu, Ma, Kong, Liu, Ji, Zhang, Chen, Chen,
  et~al.]{guo2026robustness}
Jianing Guo, Zhenhong Wu, Chang Tu, Yiyao Ma, Xiangqi Kong, Zhiqian Liu,
  Jiaming Ji, Shuning Zhang, Yuanpei Chen, Kai Chen, et~al.
\newblock On robustness of vision-language-action model against multi-modal
  perturbations.
\newblock In \emph{International Conference on Learning Representations},
  volume 2026, pp.\  70248--70272, 2026.

\bibitem[Jiang et~al.(2026)Jiang, Tan, Wheeler, Sun, Ayalew, and
  Walter]{jiang2026we}
Tianchong Jiang, Xiangshan Tan, Samuel Wheeler, Luzhe Sun, Tewodros~W Ayalew,
  and Matthew Walter.
\newblock What are we actually benchmarking in robot manipulation?
\newblock \emph{arXiv preprint arXiv:2606.04233}, 2026.

\bibitem[Kim et~al.(2024)Kim, Pertsch, Karamcheti, Xiao, Balakrishna, Nair,
  Rafailov, Foster, Lam, Sanketi, et~al.]{kim2024openvla}
Moo~Jin Kim, Karl Pertsch, Siddharth Karamcheti, Ted Xiao, Ashwin Balakrishna,
  Suraj Nair, Rafael Rafailov, Ethan Foster, Grace Lam, Pannag Sanketi, et~al.
\newblock Openvla: An open-source vision-language-action model, 2024.
\newblock \emph{URL https://arxiv. org/abs/2406.09246}, 1\penalty0
  (2):\penalty0 4, 2024.

\bibitem[Kim et~al.(2025)Kim, Finn, and Liang]{kim2025fine}
Moo~Jin Kim, Chelsea Finn, and Percy Liang.
\newblock Fine-tuning vision-language-action models: Optimizing speed and
  success.
\newblock \emph{arXiv preprint arXiv:2502.19645}, 2025.

\bibitem[Lee et~al.(2026)Lee, Kang, and Kuo]{lee2026tbd}
Sung-Wook Lee, Xuhui Kang, and Yen-Ling Kuo.
\newblock Tbd-vla: Temporal block diffusion vision language action model.
\newblock \emph{arXiv preprint arXiv:2606.07895}, 2026.

\bibitem[Li et~al.(2026)Li, Mei, Majumdar, Chen, Jiang, and
  Zhu]{li2026liberovprobenchmarkingclosedloopvisual}
Huiqiong Li, Zhiting Mei, Anirudha Majumdar, Jingjing Chen, Yu-Gang Jiang, and
  Bin Zhu.
\newblock Libero-vpro: Benchmarking closed-loop visual robustness of robotic
  foundation models, 2026.
\newblock URL \url{https://arxiv.org/abs/2609.24350}.

\bibitem[Li et~al.(2024{\natexlab{a}})Li, Liang, Wang, Luo, Chen, Liao, Wei,
  Deng, Xu, Zhang, et~al.]{li2024cogact}
Qixiu Li, Yaobo Liang, Zeyu Wang, Lin Luo, Xi~Chen, Mozheng Liao, Fangyun Wei,
  Yu~Deng, Sicheng Xu, Yizhong Zhang, et~al.
\newblock Cogact: A foundational vision-language-action model for synergizing
  cognition and action in robotic manipulation.
\newblock \emph{arXiv preprint arXiv:2411.19650}, 2024{\natexlab{a}}.

\bibitem[Li et~al.(2024{\natexlab{b}})Li, Hsu, Gu, Pertsch, Mees, Walke, Fu,
  Lunawat, Sieh, Kirmani, et~al.]{li2024evaluating}
Xuanlin Li, Kyle Hsu, Jiayuan Gu, Karl Pertsch, Oier Mees, Homer~Rich Walke,
  Chuyuan Fu, Ishikaa Lunawat, Isabel Sieh, Sean Kirmani, et~al.
\newblock Evaluating real-world robot manipulation policies in simulation.
\newblock \emph{arXiv preprint arXiv:2405.05941}, 2024{\natexlab{b}}.

\bibitem[Liu et~al.(2023)Liu, Zhu, Gao, Feng, Liu, Zhu, and
  Stone]{liu2023libero}
Bo~Liu, Yifeng Zhu, Chongkai Gao, Yihao Feng, Qiang Liu, Yuke Zhu, and Peter
  Stone.
\newblock Libero: Benchmarking knowledge transfer for lifelong robot learning.
\newblock \emph{Advances in Neural Information Processing Systems},
  36:\penalty0 44776--44791, 2023.

\bibitem[Luo et~al.(2026)Luo, Wen, Bai, Song, Liu, and Lin]{luo2026rovla}
Jingzhou Luo, Yifan Wen, Yongjie Bai, Xinshuai Song, Yang Liu, and Liang Lin.
\newblock Rovla: Multi-consistency constraints for robust
  vision-language-action models.
\newblock \emph{arXiv preprint arXiv:2605.19678}, 2026.

\bibitem[Mees et~al.(2022)Mees, Hermann, Rosete-Beas, and
  Burgard]{mees2022calvin}
Oier Mees, Lukas Hermann, Erick Rosete-Beas, and Wolfram Burgard.
\newblock Calvin: A benchmark for language-conditioned policy learning for
  long-horizon robot manipulation tasks.
\newblock \emph{IEEE Robotics and Automation Letters}, 7\penalty0 (3):\penalty0
  7327--7334, 2022.

\bibitem[Nasiriany et~al.(2024)Nasiriany, Maddukuri, Zhang, Parikh, Lo, Joshi,
  Mandlekar, and Zhu]{nasiriany2024robocasa}
Soroush Nasiriany, Abhiram Maddukuri, Lance Zhang, Adeet Parikh, Aaron Lo,
  Abhishek Joshi, Ajay Mandlekar, and Yuke Zhu.
\newblock Robocasa: Large-scale simulation of everyday tasks for generalist
  robots.
\newblock \emph{arXiv preprint arXiv:2406.02523}, 2024.

\bibitem[{NVIDIA}(2026)]{nvidia2026groot17}
{NVIDIA}.
\newblock {GR00T-N1.7-LIBERO}.
\newblock \url{https://huggingface.co/nvidia/GR00T-N1.7-LIBERO}, 2026.
\newblock Hugging Face model repository.

\bibitem[Pumacay et~al.(2024)Pumacay, Singh, Duan, Krishna, Thomason, and
  Fox]{pumacay2024colosseum}
Wilbert Pumacay, Ishika Singh, Jiafei Duan, Ranjay Krishna, Jesse Thomason, and
  Dieter Fox.
\newblock The colosseum: A benchmark for evaluating generalization for robotic
  manipulation.
\newblock \emph{arXiv preprint arXiv:2402.08191}, 2024.

\bibitem[Team et~al.(2024)Team, Ghosh, Walke, Pertsch, Black, Mees, Dasari,
  Hejna, Kreiman, Xu, et~al.]{team2024octo}
Octo~Model Team, Dibya Ghosh, Homer Walke, Karl Pertsch, Kevin Black, Oier
  Mees, Sudeep Dasari, Joey Hejna, Tobias Kreiman, Charles Xu, et~al.
\newblock Octo: An open-source generalist robot policy.
\newblock \emph{arXiv preprint arXiv:2405.12213}, 2024.

\bibitem[Xiao et~al.(2026)Xiao, Li, Gao, Ye, Jin, Qian, Zhang, Wu, and
  Yu]{xiao2026ava}
Lei Xiao, Jifeng Li, Juntao Gao, Feiyang Ye, Yan Jin, Jingjing Qian, Jing
  Zhang, Yong Wu, and Xiaoyuan Yu.
\newblock Ava-vla: improving vision-language-action models with active visual
  attention.
\newblock In \emph{Proceedings of the IEEE/CVF Conference on Computer Vision
  and Pattern Recognition}, pp.\  13453--13463, 2026.

\bibitem[Zhang et~al.(2025{\natexlab{a}})Zhang, Li, Shen, Zhang, Cai, Liu, Ji,
  Chen, Dai, Ji, et~al.]{zhang2025vla}
Borong Zhang, Jiahao Li, Jiachen Shen, Yuhao Zhang, Yishuai Cai, Lu~Liu, Hailu
  Ji, Yuanpei Chen, Juntao Dai, Jiaming Ji, et~al.
\newblock Vla-arena: An open-source framework for benchmarking
  vision-language-action models.
\newblock \emph{arXiv preprint arXiv:2512.22539}, 2025{\natexlab{a}}.

\bibitem[Zhang et~al.(2025{\natexlab{b}})Zhang, Xu, Liu, Yu, Li, Gao, Fei, Yin,
  Wu, Jiang, et~al.]{zhang2025vlabench}
Shiduo Zhang, Zhe Xu, Peiju Liu, Xiaopeng Yu, Yuan Li, Qinghui Gao, Zhaoye Fei,
  Zhangyue Yin, Zuxuan Wu, Yu-Gang Jiang, et~al.
\newblock Vlabench: A large-scale benchmark for language-conditioned robotics
  manipulation with long-horizon reasoning tasks.
\newblock In \emph{2025 IEEE/CVF International Conference on Computer Vision
  (ICCV)}, pp.\  11142--11152. IEEE, 2025{\natexlab{b}}.

\bibitem[Zhong et~al.(2026)Zhong, Liu, Wei, Xiong, Liu, and Ren]{zhong2026acot}
Linqing Zhong, Yi~Liu, Yifei Wei, Ziyu Xiong, Si~Liu, and Guanghui Ren.
\newblock Acot-vla: Action chain-of-thought for vision-language-action models.
\newblock In \emph{Proceedings of the IEEE/CVF Conference on Computer Vision
  and Pattern Recognition}, pp.\  8152--8162, 2026.

\bibitem[Zhou et~al.(2025)Zhou, Xu, Tie, Chen, Zhang, Chu, Zhou, and
  Sun]{zhou2025libero}
Xueyang Zhou, Yangming Xu, Guiyao Tie, Yongchao Chen, Guowen Zhang, Duanfeng
  Chu, Pan Zhou, and Lichao Sun.
\newblock Libero-pro: Towards robust and fair evaluation of
  vision-language-action models beyond memorization.
\newblock \emph{arXiv preprint arXiv:2510.03827}, 2025.

\end{thebibliography}
\bibliographystyle{iclr2027_conference}

\clearpage
\appendix
\section{Reproducibility and Supplementary Protocol}

This appendix documents the frozen LIBERO-MOR benchmark specification, evaluation adapters, complete aggregate results, and the severity audit. All numerical tables are generated from the episode-level JSONL files by \texttt{scripts/build\_supplement.py}; the main paper and \texttt{benchmark\_manifest.json} are not modified.

\subsection{Benchmark specification}

The benchmark contains 40 original LIBERO tasks (10 each from LIBERO-Spatial, LIBERO-Object, LIBERO-Goal, and LIBERO-10), 22 perturbation operators, three severity levels, and five seed indices. This gives 13,200 perturbed episodes and 200 paired clean references per model. The manifest is identified by SHA-256 \texttt{2b1caa800b6fd9fd6420fad15b0eef1a000919dd365327b8313ac53ba4ef2d89}. Clean and perturbed rows are paired by suite, task id, seed index, and initial-state id; the original \texttt{env.check\_success()} checker is used unchanged.

\subsection{Perturbation contract}

Table~\ref{tab:app-perturbation-parameters} lists the numerical severity contract recovered from the source implementation. The machine-readable CSV additionally records injection level, temporal scope, affected channels, and randomization rule.

\begingroup
\setlength{\tabcolsep}{2.5pt}
\renewcommand{\arraystretch}{1.08}
\begin{longtable}{@{}>{\raggedright\arraybackslash}p{1.40cm}>{\raggedright\arraybackslash}p{1.90cm}>{\raggedright\arraybackslash}p{1.30cm}>{\raggedright\arraybackslash}p{2.67cm}>{\raggedright\arraybackslash}p{2.67cm}>{\raggedright\arraybackslash}p{2.67cm}@{}}
\caption{Exact perturbation severity parameters.}\label{tab:app-perturbation-parameters}\\
\toprule
Domain & Perturbation & Injection & Severity 1 & Severity 2 & Severity 3 \\
\midrule
\endfirsthead
\multicolumn{6}{c}{\tablename\ \thetable\ (continued)}\\
\toprule
Domain & Perturbation & Injection & Severity 1 & Severity 2 & Severity 3 \\
\midrule
\endhead
\midrule
\multicolumn{6}{r}{Continued on next page}\\
\endfoot
\bottomrule
\endlastfoot
Vision & gaussian\_\allowbreak{}noise & Sensor & mean\_\allowbreak{}normalized=\allowbreak{}0.0; sigma\_\allowbreak{}normalized=\allowbreak{}0.01 & mean\_\allowbreak{}normalized=\allowbreak{}0.0; sigma\_\allowbreak{}normalized=\allowbreak{}0.03 & mean\_\allowbreak{}normalized=\allowbreak{}0.0; sigma\_\allowbreak{}normalized=\allowbreak{}0.05 \\
Vision & motion\_\allowbreak{}blur & Sensor & direction\_\allowbreak{}deg\_\allowbreak{}range=\allowbreak{}[0.0, 180.0]; kernel\_\allowbreak{}size\_\allowbreak{}px=\allowbreak{}5 & direction\_\allowbreak{}deg\_\allowbreak{}range=\allowbreak{}[0.0, 180.0]; kernel\_\allowbreak{}size\_\allowbreak{}px=\allowbreak{}9 & direction\_\allowbreak{}deg\_\allowbreak{}range=\allowbreak{}[0.0, 180.0]; kernel\_\allowbreak{}size\_\allowbreak{}px=\allowbreak{}15 \\
Vision & defocus\_\allowbreak{}blur & Sensor & gaussian\_\allowbreak{}sigma\_\allowbreak{}px=\allowbreak{}0.5 & gaussian\_\allowbreak{}sigma\_\allowbreak{}px=\allowbreak{}1.7 & gaussian\_\allowbreak{}sigma\_\allowbreak{}px=\allowbreak{}4.0 \\
Vision & compression & Sensor & codec=\allowbreak{}jpeg; quality=\allowbreak{}90 & codec=\allowbreak{}jpeg; quality=\allowbreak{}50 & codec=\allowbreak{}jpeg; quality=\allowbreak{}20 \\
Vision & brightness & Scene & light\_\allowbreak{}intensity\_\allowbreak{}delta=\allowbreak{}0.1; sampling=\allowbreak{}seeded\_\allowbreak{}dark\_\allowbreak{}or\_\allowbreak{}bright & light\_\allowbreak{}intensity\_\allowbreak{}delta=\allowbreak{}0.4; sampling=\allowbreak{}seeded\_\allowbreak{}dark\_\allowbreak{}or\_\allowbreak{}bright & light\_\allowbreak{}intensity\_\allowbreak{}delta=\allowbreak{}0.75; sampling=\allowbreak{}seeded\_\allowbreak{}dark\_\allowbreak{}or\_\allowbreak{}bright \\
Vision & shadow & Scene & ambient\_\allowbreak{}intensity\_\allowbreak{}factor=\allowbreak{}0.95; light\_\allowbreak{}direction\_\allowbreak{}jitter\_\allowbreak{}deg=\allowbreak{}3.0; light\_\allowbreak{}intensity\_\allowbreak{}factor=\allowbreak{}0.9 & ambient\_\allowbreak{}intensity\_\allowbreak{}factor=\allowbreak{}0.78; light\_\allowbreak{}direction\_\allowbreak{}jitter\_\allowbreak{}deg=\allowbreak{}12.0; light\_\allowbreak{}intensity\_\allowbreak{}factor=\allowbreak{}0.64 & ambient\_\allowbreak{}intensity\_\allowbreak{}factor=\allowbreak{}0.5; light\_\allowbreak{}direction\_\allowbreak{}jitter\_\allowbreak{}deg=\allowbreak{}30.0; light\_\allowbreak{}intensity\_\allowbreak{}factor=\allowbreak{}0.34 \\
Vision & splash & Sensor & material\_\allowbreak{}pool=\allowbreak{}[water, milk, coffee, juice]; target\_\allowbreak{}coverage\_\allowbreak{}fraction=\allowbreak{}0.05 & material\_\allowbreak{}pool=\allowbreak{}[water, milk, coffee, juice]; target\_\allowbreak{}coverage\_\allowbreak{}fraction=\allowbreak{}0.17 & material\_\allowbreak{}pool=\allowbreak{}[water, milk, coffee, juice]; target\_\allowbreak{}coverage\_\allowbreak{}fraction=\allowbreak{}0.36 \\
Vision & occlusion & Sensor & mask\_\allowbreak{}type=\allowbreak{}seeded\_\allowbreak{}polygon; target\_\allowbreak{}coverage\_\allowbreak{}fraction=\allowbreak{}0.1 & mask\_\allowbreak{}type=\allowbreak{}seeded\_\allowbreak{}polygon; target\_\allowbreak{}coverage\_\allowbreak{}fraction=\allowbreak{}0.2 & mask\_\allowbreak{}type=\allowbreak{}seeded\_\allowbreak{}polygon; target\_\allowbreak{}coverage\_\allowbreak{}fraction=\allowbreak{}0.4 \\
Vision & camera\_\allowbreak{}shake & Scene & rotation\_\allowbreak{}max\_\allowbreak{}deg=\allowbreak{}0.7; translation\_\allowbreak{}std\_\allowbreak{}m=\allowbreak{}0.002 & rotation\_\allowbreak{}max\_\allowbreak{}deg=\allowbreak{}2.8; translation\_\allowbreak{}std\_\allowbreak{}m=\allowbreak{}0.008 & rotation\_\allowbreak{}max\_\allowbreak{}deg=\allowbreak{}8.4; translation\_\allowbreak{}std\_\allowbreak{}m=\allowbreak{}0.024 \\
Vision & camera\_\allowbreak{}pose\_\allowbreak{}shift & Scene & rotation\_\allowbreak{}max\_\allowbreak{}deg=\allowbreak{}5.0; translation\_\allowbreak{}max\_\allowbreak{}m=\allowbreak{}0.01 & rotation\_\allowbreak{}max\_\allowbreak{}deg=\allowbreak{}10.0; translation\_\allowbreak{}max\_\allowbreak{}m=\allowbreak{}0.03 & rotation\_\allowbreak{}max\_\allowbreak{}deg=\allowbreak{}15.0; translation\_\allowbreak{}max\_\allowbreak{}m=\allowbreak{}0.05 \\
Language & paraphrase & Task & semantic\_\allowbreak{}constraint=\allowbreak{}preserve\_\allowbreak{}goal; synonym\_\allowbreak{}replacement\_\allowbreak{}probability=\allowbreak{}0.2 & semantic\_\allowbreak{}constraint=\allowbreak{}preserve\_\allowbreak{}goal; synonym\_\allowbreak{}replacement\_\allowbreak{}probability=\allowbreak{}0.5 & semantic\_\allowbreak{}constraint=\allowbreak{}preserve\_\allowbreak{}goal; synonym\_\allowbreak{}replacement\_\allowbreak{}probability=\allowbreak{}0.85 \\
Language & ambiguous\_\allowbreak{}instruction & Task & attribute\_\allowbreak{}removal\_\allowbreak{}probability=\allowbreak{}0.25; relation\_\allowbreak{}removal\_\allowbreak{}probability=\allowbreak{}0.25 & attribute\_\allowbreak{}removal\_\allowbreak{}probability=\allowbreak{}0.6; relation\_\allowbreak{}removal\_\allowbreak{}probability=\allowbreak{}0.6 & attribute\_\allowbreak{}removal\_\allowbreak{}probability=\allowbreak{}0.92; relation\_\allowbreak{}removal\_\allowbreak{}probability=\allowbreak{}0.92 \\
Language & distractor\_\allowbreak{}instruction & Task & num\_\allowbreak{}distractor\_\allowbreak{}clauses=\allowbreak{}1; preserve\_\allowbreak{}original\_\allowbreak{}instruction=\allowbreak{}true & num\_\allowbreak{}distractor\_\allowbreak{}clauses=\allowbreak{}2; preserve\_\allowbreak{}original\_\allowbreak{}instruction=\allowbreak{}true & num\_\allowbreak{}distractor\_\allowbreak{}clauses=\allowbreak{}3; preserve\_\allowbreak{}original\_\allowbreak{}instruction=\allowbreak{}true \\
Language & typo & Task & character\_\allowbreak{}corruption\_\allowbreak{}probability=\allowbreak{}0.03; operations=\allowbreak{}[delete, insert, substitute] & character\_\allowbreak{}corruption\_\allowbreak{}probability=\allowbreak{}0.1; operations=\allowbreak{}[delete, insert, substitute] & character\_\allowbreak{}corruption\_\allowbreak{}probability=\allowbreak{}0.24; operations=\allowbreak{}[delete, insert, substitute] \\
Proprio. & joint\_\allowbreak{}state\_\allowbreak{}noise & State & joint\_\allowbreak{}position\_\allowbreak{}std\_\allowbreak{}rad=\allowbreak{}0.01 & joint\_\allowbreak{}position\_\allowbreak{}std\_\allowbreak{}rad=\allowbreak{}0.03 & joint\_\allowbreak{}position\_\allowbreak{}std\_\allowbreak{}rad=\allowbreak{}0.05 \\
Proprio. & eef\_\allowbreak{}pose\_\allowbreak{}noise & State & position\_\allowbreak{}std\_\allowbreak{}m=\allowbreak{}0.01; rotation\_\allowbreak{}std\_\allowbreak{}deg=\allowbreak{}5.0 & position\_\allowbreak{}std\_\allowbreak{}m=\allowbreak{}0.03; rotation\_\allowbreak{}std\_\allowbreak{}deg=\allowbreak{}10.0 & position\_\allowbreak{}std\_\allowbreak{}m=\allowbreak{}0.05; rotation\_\allowbreak{}std\_\allowbreak{}deg=\allowbreak{}15.0 \\
Proprio. & missing\_\allowbreak{}Proprio. & Sensor & drop\_\allowbreak{}probability=\allowbreak{}0.1; replacement=\allowbreak{}previous\_\allowbreak{}value & drop\_\allowbreak{}probability=\allowbreak{}0.5; replacement=\allowbreak{}zero & drop\_\allowbreak{}probability=\allowbreak{}1.0; replacement=\allowbreak{}zero \\
Action & action\_\allowbreak{}execution\_\allowbreak{}noise & Actuator & action\_\allowbreak{}std\_\allowbreak{}normalized=\allowbreak{}0.01 & action\_\allowbreak{}std\_\allowbreak{}normalized=\allowbreak{}0.05 & action\_\allowbreak{}std\_\allowbreak{}normalized=\allowbreak{}0.1 \\
Action & action\_\allowbreak{}delay & Actuator & delay\_\allowbreak{}steps=\allowbreak{}1 & delay\_\allowbreak{}steps=\allowbreak{}3 & delay\_\allowbreak{}steps=\allowbreak{}5 \\
Action & action\_\allowbreak{}dropout & Actuator & drop\_\allowbreak{}probability=\allowbreak{}0.05; replacement=\allowbreak{}zero\_\allowbreak{}Action & drop\_\allowbreak{}probability=\allowbreak{}0.1; replacement=\allowbreak{}zero\_\allowbreak{}Action & drop\_\allowbreak{}probability=\allowbreak{}0.2; replacement=\allowbreak{}zero\_\allowbreak{}action \\
Env. & object\_\allowbreak{}dynamics & Scene & friction\_\allowbreak{}factor\_\allowbreak{}range=\allowbreak{}[0.8, 1.25]; mass\_\allowbreak{}factor\_\allowbreak{}range=\allowbreak{}[0.8, 1.25]; sampling=\allowbreak{}log\_\allowbreak{}uniform & friction\_\allowbreak{}factor\_\allowbreak{}range=\allowbreak{}[0.67, 1.5]; mass\_\allowbreak{}factor\_\allowbreak{}range=\allowbreak{}[0.67, 1.5]; sampling=\allowbreak{}log\_\allowbreak{}uniform & friction\_\allowbreak{}factor\_\allowbreak{}range=\allowbreak{}[0.5, 2.0]; mass\_\allowbreak{}factor\_\allowbreak{}range=\allowbreak{}[0.5, 2.0]; sampling=\allowbreak{}log\_\allowbreak{}uniform \\
Env. & distractor\_\allowbreak{}object & Scene & num\_\allowbreak{}objects=\allowbreak{}1; placement=\allowbreak{}seeded\_\allowbreak{}collision\_\allowbreak{}free & num\_\allowbreak{}objects=\allowbreak{}3; placement=\allowbreak{}seeded\_\allowbreak{}collision\_\allowbreak{}free & num\_\allowbreak{}objects=\allowbreak{}5; placement=\allowbreak{}seeded\_\allowbreak{}collision\_\allowbreak{}free \\
\end{longtable}
\endgroup

\subsection{Task inventory, pairing, and validation}

The complete task inventory is given in Table~\ref{tab:app-task-list}. The manifest contains 13,200 unique variant ids and 200 clean pairing keys. The execution audit checks row counts, duplicate keys, manifest references, execution validity, and successful invocation of the unchanged success checker.

\begingroup
\setlength{\tabcolsep}{2.5pt}
\renewcommand{\arraystretch}{1.08}
\begin{longtable}{@{}>{\raggedright\arraybackslash}p{1.50cm}>{\centering\arraybackslash}p{0.50cm}>{\raggedright\arraybackslash}p{5.30cm}>{\raggedright\arraybackslash}p{5.70cm}@{}}
\caption{The 40 original LIBERO tasks.}\label{tab:app-task-list}\\
\toprule
Suite & ID & Task & BDDL / initial-state files \\
\midrule
\endfirsthead
\multicolumn{4}{c}{\tablename\ \thetable\ (continued)}\\
\toprule
Suite & ID & Task & BDDL / initial-state files \\
\midrule
\endhead
\midrule
\multicolumn{4}{r}{Continued on next page}\\
\endfoot
\bottomrule
\endlastfoot
libero\_\allowbreak{}10 & 0 & KITCHEN SCENE3 turn on the stove and put the moka pot on it & KITCHEN\_\allowbreak{}SCENE3\_\allowbreak{}turn\_\allowbreak{}on\_\allowbreak{}the\_\allowbreak{}stove\_\allowbreak{}and\_\allowbreak{}put\_\allowbreak{}the\_\allowbreak{}moka\_\allowbreak{}pot\_\allowbreak{}on\_\allowbreak{}it.bddl / KITCHEN\_\allowbreak{}SCENE3\_\allowbreak{}turn\_\allowbreak{}on\_\allowbreak{}the\_\allowbreak{}stove\_\allowbreak{}and\_\allowbreak{}put\_\allowbreak{}the\_\allowbreak{}moka\_\allowbreak{}pot\_\allowbreak{}on\_\allowbreak{}it.pruned\_\allowbreak{}init \\
libero\_\allowbreak{}10 & 1 & KITCHEN SCENE4 put the black bowl in the bottom drawer of the cabinet and close it & KITCHEN\_\allowbreak{}SCENE4\_\allowbreak{}put\_\allowbreak{}the\_\allowbreak{}black\_\allowbreak{}bowl\_\allowbreak{}in\_\allowbreak{}the\_\allowbreak{}bottom\_\allowbreak{}drawer\_\allowbreak{}of\_\allowbreak{}the\_\allowbreak{}cabinet\_\allowbreak{}and\_\allowbreak{}close\_\allowbreak{}it.bddl / KITCHEN\_\allowbreak{}SCENE4\_\allowbreak{}put\_\allowbreak{}the\_\allowbreak{}black\_\allowbreak{}bowl\_\allowbreak{}in\_\allowbreak{}the\_\allowbreak{}bottom\_\allowbreak{}drawer\_\allowbreak{}of\_\allowbreak{}the\_\allowbreak{}cabinet\_\allowbreak{}and\_\allowbreak{}close\_\allowbreak{}it.pruned\_\allowbreak{}init \\
libero\_\allowbreak{}10 & 2 & KITCHEN SCENE6 put the yellow and white mug in the microwave and close it & KITCHEN\_\allowbreak{}SCENE6\_\allowbreak{}put\_\allowbreak{}the\_\allowbreak{}yellow\_\allowbreak{}and\_\allowbreak{}white\_\allowbreak{}mug\_\allowbreak{}in\_\allowbreak{}the\_\allowbreak{}microwave\_\allowbreak{}and\_\allowbreak{}close\_\allowbreak{}it.bddl / KITCHEN\_\allowbreak{}SCENE6\_\allowbreak{}put\_\allowbreak{}the\_\allowbreak{}yellow\_\allowbreak{}and\_\allowbreak{}white\_\allowbreak{}mug\_\allowbreak{}in\_\allowbreak{}the\_\allowbreak{}microwave\_\allowbreak{}and\_\allowbreak{}close\_\allowbreak{}it.pruned\_\allowbreak{}init \\
libero\_\allowbreak{}10 & 3 & KITCHEN SCENE8 put both moka pots on the stove & KITCHEN\_\allowbreak{}SCENE8\_\allowbreak{}put\_\allowbreak{}both\_\allowbreak{}moka\_\allowbreak{}pots\_\allowbreak{}on\_\allowbreak{}the\_\allowbreak{}stove.bddl / KITCHEN\_\allowbreak{}SCENE8\_\allowbreak{}put\_\allowbreak{}both\_\allowbreak{}moka\_\allowbreak{}pots\_\allowbreak{}on\_\allowbreak{}the\_\allowbreak{}stove.pruned\_\allowbreak{}init \\
libero\_\allowbreak{}10 & 4 & LIVING ROOM SCENE1 put both the alphabet soup and the cream cheese box in the basket & LIVING\_\allowbreak{}ROOM\_\allowbreak{}SCENE1\_\allowbreak{}put\_\allowbreak{}both\_\allowbreak{}the\_\allowbreak{}alphabet\_\allowbreak{}soup\_\allowbreak{}and\_\allowbreak{}the\_\allowbreak{}cream\_\allowbreak{}cheese\_\allowbreak{}box\_\allowbreak{}in\_\allowbreak{}the\_\allowbreak{}basket.bddl / LIVING\_\allowbreak{}ROOM\_\allowbreak{}SCENE1\_\allowbreak{}put\_\allowbreak{}both\_\allowbreak{}the\_\allowbreak{}alphabet\_\allowbreak{}soup\_\allowbreak{}and\_\allowbreak{}the\_\allowbreak{}cream\_\allowbreak{}cheese\_\allowbreak{}box\_\allowbreak{}in\_\allowbreak{}the\_\allowbreak{}basket.pruned\_\allowbreak{}init \\
libero\_\allowbreak{}10 & 5 & LIVING ROOM SCENE2 put both the alphabet soup and the tomato sauce in the basket & LIVING\_\allowbreak{}ROOM\_\allowbreak{}SCENE2\_\allowbreak{}put\_\allowbreak{}both\_\allowbreak{}the\_\allowbreak{}alphabet\_\allowbreak{}soup\_\allowbreak{}and\_\allowbreak{}the\_\allowbreak{}tomato\_\allowbreak{}sauce\_\allowbreak{}in\_\allowbreak{}the\_\allowbreak{}basket.bddl / LIVING\_\allowbreak{}ROOM\_\allowbreak{}SCENE2\_\allowbreak{}put\_\allowbreak{}both\_\allowbreak{}the\_\allowbreak{}alphabet\_\allowbreak{}soup\_\allowbreak{}and\_\allowbreak{}the\_\allowbreak{}tomato\_\allowbreak{}sauce\_\allowbreak{}in\_\allowbreak{}the\_\allowbreak{}basket.pruned\_\allowbreak{}init \\
libero\_\allowbreak{}10 & 6 & LIVING ROOM SCENE2 put both the cream cheese box and the butter in the basket & LIVING\_\allowbreak{}ROOM\_\allowbreak{}SCENE2\_\allowbreak{}put\_\allowbreak{}both\_\allowbreak{}the\_\allowbreak{}cream\_\allowbreak{}cheese\_\allowbreak{}box\_\allowbreak{}and\_\allowbreak{}the\_\allowbreak{}butter\_\allowbreak{}in\_\allowbreak{}the\_\allowbreak{}basket.bddl / LIVING\_\allowbreak{}ROOM\_\allowbreak{}SCENE2\_\allowbreak{}put\_\allowbreak{}both\_\allowbreak{}the\_\allowbreak{}cream\_\allowbreak{}cheese\_\allowbreak{}box\_\allowbreak{}and\_\allowbreak{}the\_\allowbreak{}butter\_\allowbreak{}in\_\allowbreak{}the\_\allowbreak{}basket.pruned\_\allowbreak{}init \\
libero\_\allowbreak{}10 & 7 & LIVING ROOM SCENE5 put the white mug on the left plate and put the yellow and white mug on the right plate & LIVING\_\allowbreak{}ROOM\_\allowbreak{}SCENE5\_\allowbreak{}put\_\allowbreak{}the\_\allowbreak{}white\_\allowbreak{}mug\_\allowbreak{}on\_\allowbreak{}the\_\allowbreak{}left\_\allowbreak{}plate\_\allowbreak{}and\_\allowbreak{}put\_\allowbreak{}the\_\allowbreak{}yellow\_\allowbreak{}and\_\allowbreak{}white\_\allowbreak{}mug\_\allowbreak{}on\_\allowbreak{}the\_\allowbreak{}right\_\allowbreak{}plate.bddl / LIVING\_\allowbreak{}ROOM\_\allowbreak{}SCENE5\_\allowbreak{}put\_\allowbreak{}the\_\allowbreak{}white\_\allowbreak{}mug\_\allowbreak{}on\_\allowbreak{}the\_\allowbreak{}left\_\allowbreak{}plate\_\allowbreak{}and\_\allowbreak{}put\_\allowbreak{}the\_\allowbreak{}yellow\_\allowbreak{}and\_\allowbreak{}white\_\allowbreak{}mug\_\allowbreak{}on\_\allowbreak{}the\_\allowbreak{}right\_\allowbreak{}plate.pruned\_\allowbreak{}init \\
libero\_\allowbreak{}10 & 8 & LIVING ROOM SCENE6 put the white mug on the plate and put the chocolate pudding to the right of the plate & LIVING\_\allowbreak{}ROOM\_\allowbreak{}SCENE6\_\allowbreak{}put\_\allowbreak{}the\_\allowbreak{}white\_\allowbreak{}mug\_\allowbreak{}on\_\allowbreak{}the\_\allowbreak{}plate\_\allowbreak{}and\_\allowbreak{}put\_\allowbreak{}the\_\allowbreak{}chocolate\_\allowbreak{}pudding\_\allowbreak{}to\_\allowbreak{}the\_\allowbreak{}right\_\allowbreak{}of\_\allowbreak{}the\_\allowbreak{}plate.bddl / LIVING\_\allowbreak{}ROOM\_\allowbreak{}SCENE6\_\allowbreak{}put\_\allowbreak{}the\_\allowbreak{}white\_\allowbreak{}mug\_\allowbreak{}on\_\allowbreak{}the\_\allowbreak{}plate\_\allowbreak{}and\_\allowbreak{}put\_\allowbreak{}the\_\allowbreak{}chocolate\_\allowbreak{}pudding\_\allowbreak{}to\_\allowbreak{}the\_\allowbreak{}right\_\allowbreak{}of\_\allowbreak{}the\_\allowbreak{}plate.pruned\_\allowbreak{}init \\
libero\_\allowbreak{}10 & 9 & STUDY SCENE1 pick up the book and place it in the back compartment of the caddy & STUDY\_\allowbreak{}SCENE1\_\allowbreak{}pick\_\allowbreak{}up\_\allowbreak{}the\_\allowbreak{}book\_\allowbreak{}and\_\allowbreak{}place\_\allowbreak{}it\_\allowbreak{}in\_\allowbreak{}the\_\allowbreak{}back\_\allowbreak{}compartment\_\allowbreak{}of\_\allowbreak{}the\_\allowbreak{}caddy.bddl / STUDY\_\allowbreak{}SCENE1\_\allowbreak{}pick\_\allowbreak{}up\_\allowbreak{}the\_\allowbreak{}book\_\allowbreak{}and\_\allowbreak{}place\_\allowbreak{}it\_\allowbreak{}in\_\allowbreak{}the\_\allowbreak{}back\_\allowbreak{}compartment\_\allowbreak{}of\_\allowbreak{}the\_\allowbreak{}caddy.pruned\_\allowbreak{}init \\
libero\_\allowbreak{}goal & 0 & open the middle drawer of the cabinet & open\_\allowbreak{}the\_\allowbreak{}middle\_\allowbreak{}drawer\_\allowbreak{}of\_\allowbreak{}the\_\allowbreak{}cabinet.bddl / open\_\allowbreak{}the\_\allowbreak{}middle\_\allowbreak{}drawer\_\allowbreak{}of\_\allowbreak{}the\_\allowbreak{}cabinet.pruned\_\allowbreak{}init \\
libero\_\allowbreak{}goal & 1 & open the top drawer and put the bowl inside & open\_\allowbreak{}the\_\allowbreak{}top\_\allowbreak{}drawer\_\allowbreak{}and\_\allowbreak{}put\_\allowbreak{}the\_\allowbreak{}bowl\_\allowbreak{}inside.bddl / open\_\allowbreak{}the\_\allowbreak{}top\_\allowbreak{}drawer\_\allowbreak{}and\_\allowbreak{}put\_\allowbreak{}the\_\allowbreak{}bowl\_\allowbreak{}inside.pruned\_\allowbreak{}init \\
libero\_\allowbreak{}goal & 2 & push the plate to the front of the stove & push\_\allowbreak{}the\_\allowbreak{}plate\_\allowbreak{}to\_\allowbreak{}the\_\allowbreak{}front\_\allowbreak{}of\_\allowbreak{}the\_\allowbreak{}stove.bddl / push\_\allowbreak{}the\_\allowbreak{}plate\_\allowbreak{}to\_\allowbreak{}the\_\allowbreak{}front\_\allowbreak{}of\_\allowbreak{}the\_\allowbreak{}stove.pruned\_\allowbreak{}init \\
libero\_\allowbreak{}goal & 3 & put the bowl on the plate & put\_\allowbreak{}the\_\allowbreak{}bowl\_\allowbreak{}on\_\allowbreak{}the\_\allowbreak{}plate.bddl / put\_\allowbreak{}the\_\allowbreak{}bowl\_\allowbreak{}on\_\allowbreak{}the\_\allowbreak{}plate.pruned\_\allowbreak{}init \\
libero\_\allowbreak{}goal & 4 & put the bowl on the stove & put\_\allowbreak{}the\_\allowbreak{}bowl\_\allowbreak{}on\_\allowbreak{}the\_\allowbreak{}stove.bddl / put\_\allowbreak{}the\_\allowbreak{}bowl\_\allowbreak{}on\_\allowbreak{}the\_\allowbreak{}stove.pruned\_\allowbreak{}init \\
libero\_\allowbreak{}goal & 5 & put the bowl on top of the cabinet & put\_\allowbreak{}the\_\allowbreak{}bowl\_\allowbreak{}on\_\allowbreak{}top\_\allowbreak{}of\_\allowbreak{}the\_\allowbreak{}cabinet.bddl / put\_\allowbreak{}the\_\allowbreak{}bowl\_\allowbreak{}on\_\allowbreak{}top\_\allowbreak{}of\_\allowbreak{}the\_\allowbreak{}cabinet.pruned\_\allowbreak{}init \\
libero\_\allowbreak{}goal & 6 & put the cream cheese in the bowl & put\_\allowbreak{}the\_\allowbreak{}cream\_\allowbreak{}cheese\_\allowbreak{}in\_\allowbreak{}the\_\allowbreak{}bowl.bddl / put\_\allowbreak{}the\_\allowbreak{}cream\_\allowbreak{}cheese\_\allowbreak{}in\_\allowbreak{}the\_\allowbreak{}bowl.pruned\_\allowbreak{}init \\
libero\_\allowbreak{}goal & 7 & put the wine bottle on the rack & put\_\allowbreak{}the\_\allowbreak{}wine\_\allowbreak{}bottle\_\allowbreak{}on\_\allowbreak{}the\_\allowbreak{}rack.bddl / put\_\allowbreak{}the\_\allowbreak{}wine\_\allowbreak{}bottle\_\allowbreak{}on\_\allowbreak{}the\_\allowbreak{}rack.pruned\_\allowbreak{}init \\
libero\_\allowbreak{}goal & 8 & put the wine bottle on top of the cabinet & put\_\allowbreak{}the\_\allowbreak{}wine\_\allowbreak{}bottle\_\allowbreak{}on\_\allowbreak{}top\_\allowbreak{}of\_\allowbreak{}the\_\allowbreak{}cabinet.bddl / put\_\allowbreak{}the\_\allowbreak{}wine\_\allowbreak{}bottle\_\allowbreak{}on\_\allowbreak{}top\_\allowbreak{}of\_\allowbreak{}the\_\allowbreak{}cabinet.pruned\_\allowbreak{}init \\
libero\_\allowbreak{}goal & 9 & turn on the stove & turn\_\allowbreak{}on\_\allowbreak{}the\_\allowbreak{}stove.bddl / turn\_\allowbreak{}on\_\allowbreak{}the\_\allowbreak{}stove.pruned\_\allowbreak{}init \\
libero\_\allowbreak{}object & 0 & pick up the alphabet soup and place it in the basket & pick\_\allowbreak{}up\_\allowbreak{}the\_\allowbreak{}alphabet\_\allowbreak{}soup\_\allowbreak{}and\_\allowbreak{}place\_\allowbreak{}it\_\allowbreak{}in\_\allowbreak{}the\_\allowbreak{}basket.bddl / pick\_\allowbreak{}up\_\allowbreak{}the\_\allowbreak{}alphabet\_\allowbreak{}soup\_\allowbreak{}and\_\allowbreak{}place\_\allowbreak{}it\_\allowbreak{}in\_\allowbreak{}the\_\allowbreak{}basket.pruned\_\allowbreak{}init \\
libero\_\allowbreak{}object & 1 & pick up the bbq sauce and place it in the basket & pick\_\allowbreak{}up\_\allowbreak{}the\_\allowbreak{}bbq\_\allowbreak{}sauce\_\allowbreak{}and\_\allowbreak{}place\_\allowbreak{}it\_\allowbreak{}in\_\allowbreak{}the\_\allowbreak{}basket.bddl / pick\_\allowbreak{}up\_\allowbreak{}the\_\allowbreak{}bbq\_\allowbreak{}sauce\_\allowbreak{}and\_\allowbreak{}place\_\allowbreak{}it\_\allowbreak{}in\_\allowbreak{}the\_\allowbreak{}basket.pruned\_\allowbreak{}init \\
libero\_\allowbreak{}object & 2 & pick up the butter and place it in the basket & pick\_\allowbreak{}up\_\allowbreak{}the\_\allowbreak{}butter\_\allowbreak{}and\_\allowbreak{}place\_\allowbreak{}it\_\allowbreak{}in\_\allowbreak{}the\_\allowbreak{}basket.bddl / pick\_\allowbreak{}up\_\allowbreak{}the\_\allowbreak{}butter\_\allowbreak{}and\_\allowbreak{}place\_\allowbreak{}it\_\allowbreak{}in\_\allowbreak{}the\_\allowbreak{}basket.pruned\_\allowbreak{}init \\
libero\_\allowbreak{}object & 3 & pick up the chocolate pudding and place it in the basket & pick\_\allowbreak{}up\_\allowbreak{}the\_\allowbreak{}chocolate\_\allowbreak{}pudding\_\allowbreak{}and\_\allowbreak{}place\_\allowbreak{}it\_\allowbreak{}in\_\allowbreak{}the\_\allowbreak{}basket.bddl / pick\_\allowbreak{}up\_\allowbreak{}the\_\allowbreak{}chocolate\_\allowbreak{}pudding\_\allowbreak{}and\_\allowbreak{}place\_\allowbreak{}it\_\allowbreak{}in\_\allowbreak{}the\_\allowbreak{}basket.pruned\_\allowbreak{}init \\
libero\_\allowbreak{}object & 4 & pick up the cream cheese and place it in the basket & pick\_\allowbreak{}up\_\allowbreak{}the\_\allowbreak{}cream\_\allowbreak{}cheese\_\allowbreak{}and\_\allowbreak{}place\_\allowbreak{}it\_\allowbreak{}in\_\allowbreak{}the\_\allowbreak{}basket.bddl / pick\_\allowbreak{}up\_\allowbreak{}the\_\allowbreak{}cream\_\allowbreak{}cheese\_\allowbreak{}and\_\allowbreak{}place\_\allowbreak{}it\_\allowbreak{}in\_\allowbreak{}the\_\allowbreak{}basket.pruned\_\allowbreak{}init \\
libero\_\allowbreak{}object & 5 & pick up the ketchup and place it in the basket & pick\_\allowbreak{}up\_\allowbreak{}the\_\allowbreak{}ketchup\_\allowbreak{}and\_\allowbreak{}place\_\allowbreak{}it\_\allowbreak{}in\_\allowbreak{}the\_\allowbreak{}basket.bddl / pick\_\allowbreak{}up\_\allowbreak{}the\_\allowbreak{}ketchup\_\allowbreak{}and\_\allowbreak{}place\_\allowbreak{}it\_\allowbreak{}in\_\allowbreak{}the\_\allowbreak{}basket.pruned\_\allowbreak{}init \\
libero\_\allowbreak{}object & 6 & pick up the milk and place it in the basket & pick\_\allowbreak{}up\_\allowbreak{}the\_\allowbreak{}milk\_\allowbreak{}and\_\allowbreak{}place\_\allowbreak{}it\_\allowbreak{}in\_\allowbreak{}the\_\allowbreak{}basket.bddl / pick\_\allowbreak{}up\_\allowbreak{}the\_\allowbreak{}milk\_\allowbreak{}and\_\allowbreak{}place\_\allowbreak{}it\_\allowbreak{}in\_\allowbreak{}the\_\allowbreak{}basket.pruned\_\allowbreak{}init \\
libero\_\allowbreak{}object & 7 & pick up the orange juice and place it in the basket & pick\_\allowbreak{}up\_\allowbreak{}the\_\allowbreak{}orange\_\allowbreak{}juice\_\allowbreak{}and\_\allowbreak{}place\_\allowbreak{}it\_\allowbreak{}in\_\allowbreak{}the\_\allowbreak{}basket.bddl / pick\_\allowbreak{}up\_\allowbreak{}the\_\allowbreak{}orange\_\allowbreak{}juice\_\allowbreak{}and\_\allowbreak{}place\_\allowbreak{}it\_\allowbreak{}in\_\allowbreak{}the\_\allowbreak{}basket.pruned\_\allowbreak{}init \\
libero\_\allowbreak{}object & 8 & pick up the salad dressing and place it in the basket & pick\_\allowbreak{}up\_\allowbreak{}the\_\allowbreak{}salad\_\allowbreak{}dressing\_\allowbreak{}and\_\allowbreak{}place\_\allowbreak{}it\_\allowbreak{}in\_\allowbreak{}the\_\allowbreak{}basket.bddl / pick\_\allowbreak{}up\_\allowbreak{}the\_\allowbreak{}salad\_\allowbreak{}dressing\_\allowbreak{}and\_\allowbreak{}place\_\allowbreak{}it\_\allowbreak{}in\_\allowbreak{}the\_\allowbreak{}basket.pruned\_\allowbreak{}init \\
libero\_\allowbreak{}object & 9 & pick up the tomato sauce and place it in the basket & pick\_\allowbreak{}up\_\allowbreak{}the\_\allowbreak{}tomato\_\allowbreak{}sauce\_\allowbreak{}and\_\allowbreak{}place\_\allowbreak{}it\_\allowbreak{}in\_\allowbreak{}the\_\allowbreak{}basket.bddl / pick\_\allowbreak{}up\_\allowbreak{}the\_\allowbreak{}tomato\_\allowbreak{}sauce\_\allowbreak{}and\_\allowbreak{}place\_\allowbreak{}it\_\allowbreak{}in\_\allowbreak{}the\_\allowbreak{}basket.pruned\_\allowbreak{}init \\
libero\_\allowbreak{}spatial & 0 & pick up the black bowl between the plate and the ramekin and place it on the plate & pick\_\allowbreak{}up\_\allowbreak{}the\_\allowbreak{}black\_\allowbreak{}bowl\_\allowbreak{}between\_\allowbreak{}the\_\allowbreak{}plate\_\allowbreak{}and\_\allowbreak{}the\_\allowbreak{}ramekin\_\allowbreak{}and\_\allowbreak{}place\_\allowbreak{}it\_\allowbreak{}on\_\allowbreak{}the\_\allowbreak{}plate.bddl / pick\_\allowbreak{}up\_\allowbreak{}the\_\allowbreak{}black\_\allowbreak{}bowl\_\allowbreak{}between\_\allowbreak{}the\_\allowbreak{}plate\_\allowbreak{}and\_\allowbreak{}the\_\allowbreak{}ramekin\_\allowbreak{}and\_\allowbreak{}place\_\allowbreak{}it\_\allowbreak{}on\_\allowbreak{}the\_\allowbreak{}plate.pruned\_\allowbreak{}init \\
libero\_\allowbreak{}spatial & 1 & pick up the black bowl from table center and place it on the plate & pick\_\allowbreak{}up\_\allowbreak{}the\_\allowbreak{}black\_\allowbreak{}bowl\_\allowbreak{}from\_\allowbreak{}table\_\allowbreak{}center\_\allowbreak{}and\_\allowbreak{}place\_\allowbreak{}it\_\allowbreak{}on\_\allowbreak{}the\_\allowbreak{}plate.bddl / pick\_\allowbreak{}up\_\allowbreak{}the\_\allowbreak{}black\_\allowbreak{}bowl\_\allowbreak{}from\_\allowbreak{}table\_\allowbreak{}center\_\allowbreak{}and\_\allowbreak{}place\_\allowbreak{}it\_\allowbreak{}on\_\allowbreak{}the\_\allowbreak{}plate.pruned\_\allowbreak{}init \\
libero\_\allowbreak{}spatial & 2 & pick up the black bowl in the top drawer of the wooden cabinet and place it on the plate & pick\_\allowbreak{}up\_\allowbreak{}the\_\allowbreak{}black\_\allowbreak{}bowl\_\allowbreak{}in\_\allowbreak{}the\_\allowbreak{}top\_\allowbreak{}drawer\_\allowbreak{}of\_\allowbreak{}the\_\allowbreak{}wooden\_\allowbreak{}cabinet\_\allowbreak{}and\_\allowbreak{}place\_\allowbreak{}it\_\allowbreak{}on\_\allowbreak{}the\_\allowbreak{}plate.bddl / pick\_\allowbreak{}up\_\allowbreak{}the\_\allowbreak{}black\_\allowbreak{}bowl\_\allowbreak{}in\_\allowbreak{}the\_\allowbreak{}top\_\allowbreak{}drawer\_\allowbreak{}of\_\allowbreak{}the\_\allowbreak{}wooden\_\allowbreak{}cabinet\_\allowbreak{}and\_\allowbreak{}place\_\allowbreak{}it\_\allowbreak{}on\_\allowbreak{}the\_\allowbreak{}plate.pruned\_\allowbreak{}init \\
libero\_\allowbreak{}spatial & 3 & pick up the black bowl next to the cookie box and place it on the plate & pick\_\allowbreak{}up\_\allowbreak{}the\_\allowbreak{}black\_\allowbreak{}bowl\_\allowbreak{}next\_\allowbreak{}to\_\allowbreak{}the\_\allowbreak{}cookie\_\allowbreak{}box\_\allowbreak{}and\_\allowbreak{}place\_\allowbreak{}it\_\allowbreak{}on\_\allowbreak{}the\_\allowbreak{}plate.bddl / pick\_\allowbreak{}up\_\allowbreak{}the\_\allowbreak{}black\_\allowbreak{}bowl\_\allowbreak{}next\_\allowbreak{}to\_\allowbreak{}the\_\allowbreak{}cookie\_\allowbreak{}box\_\allowbreak{}and\_\allowbreak{}place\_\allowbreak{}it\_\allowbreak{}on\_\allowbreak{}the\_\allowbreak{}plate.pruned\_\allowbreak{}init \\
libero\_\allowbreak{}spatial & 4 & pick up the black bowl next to the plate and place it on the plate & pick\_\allowbreak{}up\_\allowbreak{}the\_\allowbreak{}black\_\allowbreak{}bowl\_\allowbreak{}next\_\allowbreak{}to\_\allowbreak{}the\_\allowbreak{}plate\_\allowbreak{}and\_\allowbreak{}place\_\allowbreak{}it\_\allowbreak{}on\_\allowbreak{}the\_\allowbreak{}plate.bddl / pick\_\allowbreak{}up\_\allowbreak{}the\_\allowbreak{}black\_\allowbreak{}bowl\_\allowbreak{}next\_\allowbreak{}to\_\allowbreak{}the\_\allowbreak{}plate\_\allowbreak{}and\_\allowbreak{}place\_\allowbreak{}it\_\allowbreak{}on\_\allowbreak{}the\_\allowbreak{}plate.pruned\_\allowbreak{}init \\
libero\_\allowbreak{}spatial & 5 & pick up the black bowl next to the ramekin and place it on the plate & pick\_\allowbreak{}up\_\allowbreak{}the\_\allowbreak{}black\_\allowbreak{}bowl\_\allowbreak{}next\_\allowbreak{}to\_\allowbreak{}the\_\allowbreak{}ramekin\_\allowbreak{}and\_\allowbreak{}place\_\allowbreak{}it\_\allowbreak{}on\_\allowbreak{}the\_\allowbreak{}plate.bddl / pick\_\allowbreak{}up\_\allowbreak{}the\_\allowbreak{}black\_\allowbreak{}bowl\_\allowbreak{}next\_\allowbreak{}to\_\allowbreak{}the\_\allowbreak{}ramekin\_\allowbreak{}and\_\allowbreak{}place\_\allowbreak{}it\_\allowbreak{}on\_\allowbreak{}the\_\allowbreak{}plate.pruned\_\allowbreak{}init \\
libero\_\allowbreak{}spatial & 6 & pick up the black bowl on the cookie box and place it on the plate & pick\_\allowbreak{}up\_\allowbreak{}the\_\allowbreak{}black\_\allowbreak{}bowl\_\allowbreak{}on\_\allowbreak{}the\_\allowbreak{}cookie\_\allowbreak{}box\_\allowbreak{}and\_\allowbreak{}place\_\allowbreak{}it\_\allowbreak{}on\_\allowbreak{}the\_\allowbreak{}plate.bddl / pick\_\allowbreak{}up\_\allowbreak{}the\_\allowbreak{}black\_\allowbreak{}bowl\_\allowbreak{}on\_\allowbreak{}the\_\allowbreak{}cookie\_\allowbreak{}box\_\allowbreak{}and\_\allowbreak{}place\_\allowbreak{}it\_\allowbreak{}on\_\allowbreak{}the\_\allowbreak{}plate.pruned\_\allowbreak{}init \\
libero\_\allowbreak{}spatial & 7 & pick up the black bowl on the ramekin and place it on the plate & pick\_\allowbreak{}up\_\allowbreak{}the\_\allowbreak{}black\_\allowbreak{}bowl\_\allowbreak{}on\_\allowbreak{}the\_\allowbreak{}ramekin\_\allowbreak{}and\_\allowbreak{}place\_\allowbreak{}it\_\allowbreak{}on\_\allowbreak{}the\_\allowbreak{}plate.bddl / pick\_\allowbreak{}up\_\allowbreak{}the\_\allowbreak{}black\_\allowbreak{}bowl\_\allowbreak{}on\_\allowbreak{}the\_\allowbreak{}ramekin\_\allowbreak{}and\_\allowbreak{}place\_\allowbreak{}it\_\allowbreak{}on\_\allowbreak{}the\_\allowbreak{}plate.pruned\_\allowbreak{}init \\
libero\_\allowbreak{}spatial & 8 & pick up the black bowl on the stove and place it on the plate & pick\_\allowbreak{}up\_\allowbreak{}the\_\allowbreak{}black\_\allowbreak{}bowl\_\allowbreak{}on\_\allowbreak{}the\_\allowbreak{}stove\_\allowbreak{}and\_\allowbreak{}place\_\allowbreak{}it\_\allowbreak{}on\_\allowbreak{}the\_\allowbreak{}plate.bddl / pick\_\allowbreak{}up\_\allowbreak{}the\_\allowbreak{}black\_\allowbreak{}bowl\_\allowbreak{}on\_\allowbreak{}the\_\allowbreak{}stove\_\allowbreak{}and\_\allowbreak{}place\_\allowbreak{}it\_\allowbreak{}on\_\allowbreak{}the\_\allowbreak{}plate.pruned\_\allowbreak{}init \\
libero\_\allowbreak{}spatial & 9 & pick up the black bowl on the wooden cabinet and place it on the plate & pick\_\allowbreak{}up\_\allowbreak{}the\_\allowbreak{}black\_\allowbreak{}bowl\_\allowbreak{}on\_\allowbreak{}the\_\allowbreak{}wooden\_\allowbreak{}cabinet\_\allowbreak{}and\_\allowbreak{}place\_\allowbreak{}it\_\allowbreak{}on\_\allowbreak{}the\_\allowbreak{}plate.bddl / pick\_\allowbreak{}up\_\allowbreak{}the\_\allowbreak{}black\_\allowbreak{}bowl\_\allowbreak{}on\_\allowbreak{}the\_\allowbreak{}wooden\_\allowbreak{}cabinet\_\allowbreak{}and\_\allowbreak{}place\_\allowbreak{}it\_\allowbreak{}on\_\allowbreak{}the\_\allowbreak{}plate.pruned\_\allowbreak{}init \\
\end{longtable}
\endgroup

\clearpage
\subsection{Models and adapters}

Table~\ref{tab:app-model-configs} records the checkpoint identifiers
and adapter entry points available in the frozen evaluation metadata.
Metadata that were not captured by the frozen rollout records are
omitted rather than inferred.

\begin{table}[H]
\centering
\caption{Resolved model checkpoints and adapter contracts.}
\label{tab:app-model-configs}
\begingroup
\setlength{\tabcolsep}{3pt}
\renewcommand{\arraystretch}{1.10}
\begin{tabularx}{\linewidth}{@{}>{\raggedright\arraybackslash}p{2.35cm}>{\raggedright\arraybackslash}p{4.15cm}>{\raggedright\arraybackslash}p{4.25cm}>{\raggedright\arraybackslash}X@{}}
\toprule
Model & Checkpoint & Adapter & Action contract \\
\midrule
OpenVLA-\allowbreak{}OFT & openvla-\allowbreak{}7b-\allowbreak{}oft-\allowbreak{}finetuned-\allowbreak{}libero-\allowbreak{}spatial-\allowbreak{}object-\allowbreak{}goal-\allowbreak{}10 & libero/\allowbreak{}libero/\allowbreak{}benchmark/\allowbreak{}models/\allowbreak{}openvla\_\allowbreak{}oft\_\allowbreak{}adapter.py & continuous 7D, clipped to [-\allowbreak{}1,1] by adapter/\allowbreak{}controller \\
pi0.5 & pi05\_\allowbreak{}libero\_\allowbreak{}pytorch\_\allowbreak{}retry2 & libero/\allowbreak{}libero/\allowbreak{}benchmark/\allowbreak{}models/\allowbreak{}pi0\_\allowbreak{}adapter.py & flow-\allowbreak{}matching 7D action; benchmark executes 10 steps before replanning \\
UniVLA & univla-\allowbreak{}7b-\allowbreak{}224-\allowbreak{}sft-\allowbreak{}libero & libero/\allowbreak{}libero/\allowbreak{}benchmark/\allowbreak{}models/\allowbreak{}univla\_\allowbreak{}adapter.py & decoded 7D action, gripper conversion, clip [-\allowbreak{}1,1] \\
GR00T-\allowbreak{}N1.7-\allowbreak{}LIBERO & GR00T-\allowbreak{}N1.7-\allowbreak{}LIBERO & benchmark\_\allowbreak{}plugins/\allowbreak{}rpc.py + GR00T backend configuration & continuous 7D delta action \\
AVA-\allowbreak{}VLA & avavla-\allowbreak{}libero-\allowbreak{}4in1 & benchmark\_\allowbreak{}plugins/\allowbreak{}avavla.py & 8-\allowbreak{}action chunk; OXE gripper conversion to LIBERO 7D \\
TBD-\allowbreak{}VLA & tbdvla\_\allowbreak{}libero & benchmark\_\allowbreak{}plugins/\allowbreak{}tbdvla.py & 7D output with saved unnormalizer and LIBERO gripper convention \\
CogACT-\allowbreak{}Libero & CogACT-\allowbreak{}Libero & libero/\allowbreak{}libero/\allowbreak{}benchmark/\allowbreak{}models/\allowbreak{}cogact\_\allowbreak{}adapter.py & OXE action converted to LIBERO 7D; native signed gripper \\
\bottomrule\end{tabularx}
\endgroup

\end{table}
\FloatBarrier

\subsection{Complete results}

The overall results in Table~\ref{tab:app-overall} exactly reproduce
the frozen main-paper values after filtering the shared
OpenVLA-OFT/pi0.5 result files by policy name. Domain and suite
breakdowns are provided in machine-readable CSV files and compact
appendix tables; the complete model--perturbation and model--severity
matrices are in
\texttt{artifacts/results\_by\_perturbation.csv} and
\texttt{artifacts/results\_by\_severity.csv}.

\begin{table}[H]
\centering
\caption{Overall clean and perturbed success rates.}
\label{tab:app-overall}
\begin{tabular}{lrrrr}\toprule
Model & Clean SR & Perturbed SR & $\Delta$SR & RR \\
\midrule
OpenVLA-\allowbreak{}OFT & 92.50 & 48.64 & -43.86 & 52.58 \\
pi0.5 & 97.00 & 62.68 & -34.32 & 64.62 \\
UniVLA & 79.00 & 39.68 & -39.32 & 50.23 \\
GR00T-\allowbreak{}N1.7-\allowbreak{}LIBERO & 95.00 & 57.38 & -37.62 & 60.40 \\
AVA-\allowbreak{}VLA & 95.00 & 49.67 & -45.33 & 52.29 \\
TBD-\allowbreak{}VLA & 95.50 & 59.50 & -36.00 & 62.30 \\
CogACT-\allowbreak{}Libero & 94.00 & 58.33 & -35.67 & 62.06 \\
\bottomrule\end{tabular}

\end{table}
\FloatBarrier

\begingroup
\setlength{\tabcolsep}{2.5pt}
\renewcommand{\arraystretch}{1.08}
\begin{longtable}{@{}>{\raggedright\arraybackslash}p{3.7cm}>{\raggedright\arraybackslash}p{3.0cm}rr@{}}
\caption{Perturbed success rates by modality.}\label{tab:app-domain}\\
\toprule
Model & Domain & Perturbed SR & RR \\
\midrule
\endfirsthead
\multicolumn{4}{c}{\tablename\ \thetable\ (continued)}\\
\toprule
Model & Domain & Perturbed SR & RR \\
\midrule
\endhead
\midrule
\multicolumn{4}{r}{Continued on next page}\\
\endfoot
\bottomrule
\endlastfoot
OpenVLA-\allowbreak{}OFT & vision & 47.60 & 51.46 \\
OpenVLA-\allowbreak{}OFT & language & 45.38 & 49.05 \\
OpenVLA-\allowbreak{}OFT & proprioception & 41.67 & 45.05 \\
OpenVLA-\allowbreak{}OFT & action & 54.39 & 58.80 \\
OpenVLA-\allowbreak{}OFT & environment & 62.17 & 67.21 \\
pi0.5 & vision & 56.23 & 57.97 \\
pi0.5 & language & 66.71 & 68.77 \\
pi0.5 & proprioception & 64.72 & 66.72 \\
pi0.5 & action & 67.61 & 69.70 \\
pi0.5 & environment & 76.42 & 78.78 \\
UniVLA & vision & 31.20 & 39.49 \\
UniVLA & language & 45.96 & 58.18 \\
UniVLA & proprioception & 44.00 & 55.70 \\
UniVLA & action & 45.11 & 57.10 \\
UniVLA & environment & 54.92 & 69.51 \\
GR00T-\allowbreak{}N1.7-\allowbreak{}LIBERO & vision & 57.05 & 60.05 \\
GR00T-\allowbreak{}N1.7-\allowbreak{}LIBERO & language & 54.96 & 57.85 \\
GR00T-\allowbreak{}N1.7-\allowbreak{}LIBERO & proprioception & 53.50 & 56.32 \\
GR00T-\allowbreak{}N1.7-\allowbreak{}LIBERO & action & 55.78 & 58.71 \\
GR00T-\allowbreak{}N1.7-\allowbreak{}LIBERO & environment & 72.08 & 75.88 \\
AVA-\allowbreak{}VLA & vision & 51.15 & 53.84 \\
AVA-\allowbreak{}VLA & language & 43.75 & 46.05 \\
AVA-\allowbreak{}VLA & proprioception & 39.78 & 41.87 \\
AVA-\allowbreak{}VLA & action & 52.39 & 55.15 \\
AVA-\allowbreak{}VLA & environment & 64.92 & 68.33 \\
TBD-\allowbreak{}VLA & vision & 60.85 & 63.72 \\
TBD-\allowbreak{}VLA & language & 53.17 & 55.67 \\
TBD-\allowbreak{}VLA & proprioception & 49.72 & 52.07 \\
TBD-\allowbreak{}VLA & action & 62.83 & 65.79 \\
TBD-\allowbreak{}VLA & environment & 75.08 & 78.62 \\
CogACT-\allowbreak{}Libero & vision & 57.32 & 60.98 \\
CogACT-\allowbreak{}Libero & language & 55.21 & 58.73 \\
CogACT-\allowbreak{}Libero & proprioception & 56.83 & 60.46 \\
CogACT-\allowbreak{}Libero & action & 57.06 & 60.70 \\
CogACT-\allowbreak{}Libero & environment & 73.83 & 78.55 \\
\end{longtable}
\endgroup

\subsection{Severity and non-monotonicity audit}

For every model--perturbation pair, we report success rates at severity levels 1--3 and paired adjacent-level transitions. A curve is flagged when the point estimate increases from one severity to the next. This is an empirical property of policy/environment trajectories and is not treated as evidence that the ordered operator contract is invalid. The full transition table identifies affected task--seed keys, task names, and seed indices.

\begingroup
\setlength{\tabcolsep}{2.5pt}
\renewcommand{\arraystretch}{1.08}
\begin{longtable}{@{}>{\raggedright\arraybackslash}p{3.0cm}>{\raggedright\arraybackslash}p{4.0cm}rrr>{\raggedright\arraybackslash}p{2.0cm}@{}}
\caption{Severity curves and flagged reversals.}\label{tab:app-nonmonotonicity}\\
\toprule
Model & Perturbation & S1 & S2 & S3 & Reversals \\
\midrule
\endfirsthead
\multicolumn{6}{c}{\tablename\ \thetable\ (continued)}\\
\toprule
Model & Perturbation & S1 & S2 & S3 & Reversals \\
\midrule
\endhead
\midrule
\multicolumn{6}{r}{Continued on next page}\\
\endfoot
\bottomrule
\endlastfoot
AVA-\allowbreak{}VLA & action\_\allowbreak{}delay & 61.50 & 40.50 & 23.00 & none \\
AVA-\allowbreak{}VLA & action\_\allowbreak{}dropout & 62.50 & 64.00 & 60.50 & 1-\allowbreak{}>2 \\
AVA-\allowbreak{}VLA & action\_\allowbreak{}execution\_\allowbreak{}noise & 57.00 & 54.00 & 48.50 & none \\
AVA-\allowbreak{}VLA & ambiguous\_\allowbreak{}instruction & 52.50 & 52.50 & 56.50 & 2-\allowbreak{}>3 \\
AVA-\allowbreak{}VLA & brightness & 58.00 & 63.00 & 59.00 & 1-\allowbreak{}>2 \\
AVA-\allowbreak{}VLA & camera\_\allowbreak{}pose\_\allowbreak{}shift & 61.00 & 48.00 & 25.00 & none \\
AVA-\allowbreak{}VLA & camera\_\allowbreak{}shake & 63.00 & 56.50 & 27.00 & none \\
AVA-\allowbreak{}VLA & compression & 61.00 & 63.50 & 61.00 & 1-\allowbreak{}>2 \\
AVA-\allowbreak{}VLA & defocus\_\allowbreak{}blur & 64.50 & 61.00 & 27.50 & none \\
AVA-\allowbreak{}VLA & distractor\_\allowbreak{}instruction & 1.50 & 1.50 & 1.00 & none \\
AVA-\allowbreak{}VLA & distractor\_\allowbreak{}object & 67.50 & 66.00 & 70.50 & 2-\allowbreak{}>3 \\
AVA-\allowbreak{}VLA & eef\_\allowbreak{}pose\_\allowbreak{}noise & 49.50 & 22.00 & 12.00 & none \\
AVA-\allowbreak{}VLA & gaussian\_\allowbreak{}noise & 64.50 & 64.50 & 60.00 & none \\
AVA-\allowbreak{}VLA & joint\_\allowbreak{}state\_\allowbreak{}noise & 63.50 & 54.50 & 51.00 & none \\
AVA-\allowbreak{}VLA & missing\_\allowbreak{}proprioception & 61.00 & 39.50 & 5.00 & none \\
AVA-\allowbreak{}VLA & motion\_\allowbreak{}blur & 62.00 & 60.00 & 39.00 & none \\
AVA-\allowbreak{}VLA & object\_\allowbreak{}dynamics & 62.50 & 63.50 & 59.50 & 1-\allowbreak{}>2 \\
AVA-\allowbreak{}VLA & occlusion & 45.00 & 20.50 & 11.50 & none \\
AVA-\allowbreak{}VLA & paraphrase & 61.00 & 64.00 & 61.00 & 1-\allowbreak{}>2 \\
AVA-\allowbreak{}VLA & shadow & 61.50 & 61.00 & 61.50 & 2-\allowbreak{}>3 \\
AVA-\allowbreak{}VLA & splash & 52.00 & 42.50 & 30.00 & none \\
AVA-\allowbreak{}VLA & typo & 61.50 & 59.50 & 52.50 & none \\
CogACT-\allowbreak{}Libero & action\_\allowbreak{}delay & 62.50 & 44.00 & 22.50 & none \\
CogACT-\allowbreak{}Libero & action\_\allowbreak{}dropout & 68.50 & 64.00 & 63.50 & none \\
CogACT-\allowbreak{}Libero & action\_\allowbreak{}execution\_\allowbreak{}noise & 64.00 & 66.00 & 58.50 & 1-\allowbreak{}>2 \\
CogACT-\allowbreak{}Libero & ambiguous\_\allowbreak{}instruction & 40.50 & 37.50 & 40.00 & 2-\allowbreak{}>3 \\
CogACT-\allowbreak{}Libero & brightness & 64.00 & 64.00 & 62.00 & none \\
CogACT-\allowbreak{}Libero & camera\_\allowbreak{}pose\_\allowbreak{}shift & 63.50 & 50.50 & 39.50 & none \\
CogACT-\allowbreak{}Libero & camera\_\allowbreak{}shake & 62.50 & 66.50 & 50.00 & 1-\allowbreak{}>2 \\
CogACT-\allowbreak{}Libero & compression & 67.00 & 63.50 & 66.50 & 2-\allowbreak{}>3 \\
CogACT-\allowbreak{}Libero & defocus\_\allowbreak{}blur & 64.50 & 67.50 & 40.50 & 1-\allowbreak{}>2 \\
CogACT-\allowbreak{}Libero & distractor\_\allowbreak{}instruction & 64.50 & 65.50 & 60.50 & 1-\allowbreak{}>2 \\
CogACT-\allowbreak{}Libero & distractor\_\allowbreak{}object & 72.50 & 76.00 & 87.50 & 1-\allowbreak{}>2,2-\allowbreak{}>3 \\
CogACT-\allowbreak{}Libero & eef\_\allowbreak{}pose\_\allowbreak{}noise & 58.50 & 38.50 & 19.50 & none \\
CogACT-\allowbreak{}Libero & gaussian\_\allowbreak{}noise & 67.50 & 65.00 & 65.00 & none \\
CogACT-\allowbreak{}Libero & joint\_\allowbreak{}state\_\allowbreak{}noise & 67.50 & 66.50 & 61.00 & none \\
CogACT-\allowbreak{}Libero & missing\_\allowbreak{}proprioception & 67.50 & 65.50 & 67.00 & 2-\allowbreak{}>3 \\
CogACT-\allowbreak{}Libero & motion\_\allowbreak{}blur & 65.50 & 63.50 & 58.00 & none \\
CogACT-\allowbreak{}Libero & object\_\allowbreak{}dynamics & 68.50 & 69.00 & 69.50 & 1-\allowbreak{}>2,2-\allowbreak{}>3 \\
CogACT-\allowbreak{}Libero & occlusion & 45.00 & 23.50 & 15.50 & none \\
CogACT-\allowbreak{}Libero & paraphrase & 63.50 & 64.00 & 68.50 & 1-\allowbreak{}>2,2-\allowbreak{}>3 \\
CogACT-\allowbreak{}Libero & shadow & 68.00 & 66.00 & 65.50 & none \\
CogACT-\allowbreak{}Libero & splash & 58.50 & 55.00 & 46.00 & none \\
CogACT-\allowbreak{}Libero & typo & 64.50 & 56.50 & 37.00 & none \\
GR00T-\allowbreak{}N1.7-\allowbreak{}LIBERO & action\_\allowbreak{}delay & 60.00 & 45.00 & 21.00 & none \\
GR00T-\allowbreak{}N1.7-\allowbreak{}LIBERO & action\_\allowbreak{}dropout & 68.50 & 65.50 & 63.00 & none \\
GR00T-\allowbreak{}N1.7-\allowbreak{}LIBERO & action\_\allowbreak{}execution\_\allowbreak{}noise & 62.50 & 61.00 & 55.50 & none \\
GR00T-\allowbreak{}N1.7-\allowbreak{}LIBERO & ambiguous\_\allowbreak{}instruction & 44.00 & 42.00 & 40.50 & none \\
GR00T-\allowbreak{}N1.7-\allowbreak{}LIBERO & brightness & 62.00 & 66.50 & 61.00 & 1-\allowbreak{}>2 \\
GR00T-\allowbreak{}N1.7-\allowbreak{}LIBERO & camera\_\allowbreak{}pose\_\allowbreak{}shift & 63.00 & 46.00 & 29.00 & none \\
GR00T-\allowbreak{}N1.7-\allowbreak{}LIBERO & camera\_\allowbreak{}shake & 66.00 & 64.00 & 31.50 & none \\
GR00T-\allowbreak{}N1.7-\allowbreak{}LIBERO & compression & 61.50 & 67.00 & 65.00 & 1-\allowbreak{}>2 \\
GR00T-\allowbreak{}N1.7-\allowbreak{}LIBERO & defocus\_\allowbreak{}blur & 67.00 & 64.00 & 47.00 & none \\
GR00T-\allowbreak{}N1.7-\allowbreak{}LIBERO & distractor\_\allowbreak{}instruction & 63.00 & 64.50 & 61.50 & 1-\allowbreak{}>2 \\
GR00T-\allowbreak{}N1.7-\allowbreak{}LIBERO & distractor\_\allowbreak{}object & 72.00 & 78.50 & 81.50 & 1-\allowbreak{}>2,2-\allowbreak{}>3 \\
GR00T-\allowbreak{}N1.7-\allowbreak{}LIBERO & eef\_\allowbreak{}pose\_\allowbreak{}noise & 59.50 & 33.50 & 16.50 & none \\
GR00T-\allowbreak{}N1.7-\allowbreak{}LIBERO & gaussian\_\allowbreak{}noise & 66.00 & 66.00 & 63.50 & none \\
GR00T-\allowbreak{}N1.7-\allowbreak{}LIBERO & joint\_\allowbreak{}state\_\allowbreak{}noise & 66.00 & 60.00 & 60.50 & 2-\allowbreak{}>3 \\
GR00T-\allowbreak{}N1.7-\allowbreak{}LIBERO & missing\_\allowbreak{}proprioception & 59.50 & 61.50 & 64.50 & 1-\allowbreak{}>2,2-\allowbreak{}>3 \\
GR00T-\allowbreak{}N1.7-\allowbreak{}LIBERO & motion\_\allowbreak{}blur & 63.00 & 63.50 & 53.00 & 1-\allowbreak{}>2 \\
GR00T-\allowbreak{}N1.7-\allowbreak{}LIBERO & object\_\allowbreak{}dynamics & 67.50 & 66.50 & 66.50 & none \\
GR00T-\allowbreak{}N1.7-\allowbreak{}LIBERO & occlusion & 57.00 & 41.50 & 20.50 & none \\
GR00T-\allowbreak{}N1.7-\allowbreak{}LIBERO & paraphrase & 61.50 & 62.00 & 61.00 & 1-\allowbreak{}>2 \\
GR00T-\allowbreak{}N1.7-\allowbreak{}LIBERO & shadow & 63.00 & 63.00 & 62.50 & none \\
GR00T-\allowbreak{}N1.7-\allowbreak{}LIBERO & splash & 58.50 & 55.00 & 55.00 & none \\
GR00T-\allowbreak{}N1.7-\allowbreak{}LIBERO & typo & 61.00 & 56.00 & 42.50 & none \\
OpenVLA-\allowbreak{}OFT & action\_\allowbreak{}delay & 59.00 & 44.00 & 14.50 & none \\
OpenVLA-\allowbreak{}OFT & action\_\allowbreak{}dropout & 64.50 & 70.00 & 64.00 & 1-\allowbreak{}>2 \\
OpenVLA-\allowbreak{}OFT & action\_\allowbreak{}execution\_\allowbreak{}noise & 63.50 & 59.00 & 51.00 & none \\
OpenVLA-\allowbreak{}OFT & ambiguous\_\allowbreak{}instruction & 59.50 & 59.00 & 61.00 & 2-\allowbreak{}>3 \\
OpenVLA-\allowbreak{}OFT & brightness & 68.50 & 67.00 & 65.00 & none \\
OpenVLA-\allowbreak{}OFT & camera\_\allowbreak{}pose\_\allowbreak{}shift & 58.50 & 43.00 & 29.50 & none \\
OpenVLA-\allowbreak{}OFT & camera\_\allowbreak{}shake & 66.50 & 59.50 & 25.00 & none \\
OpenVLA-\allowbreak{}OFT & compression & 70.00 & 62.00 & 63.00 & 2-\allowbreak{}>3 \\
OpenVLA-\allowbreak{}OFT & defocus\_\allowbreak{}blur & 68.00 & 37.00 & 1.50 & none \\
OpenVLA-\allowbreak{}OFT & distractor\_\allowbreak{}instruction & 0.50 & 0.50 & 0.00 & none \\
OpenVLA-\allowbreak{}OFT & distractor\_\allowbreak{}object & 64.50 & 61.00 & 64.00 & 2-\allowbreak{}>3 \\
OpenVLA-\allowbreak{}OFT & eef\_\allowbreak{}pose\_\allowbreak{}noise & 52.00 & 24.00 & 11.50 & none \\
OpenVLA-\allowbreak{}OFT & gaussian\_\allowbreak{}noise & 64.50 & 57.50 & 56.00 & none \\
OpenVLA-\allowbreak{}OFT & joint\_\allowbreak{}state\_\allowbreak{}noise & 67.50 & 62.00 & 57.00 & none \\
OpenVLA-\allowbreak{}OFT & missing\_\allowbreak{}proprioception & 61.00 & 33.50 & 6.50 & none \\
OpenVLA-\allowbreak{}OFT & motion\_\allowbreak{}blur & 56.00 & 26.50 & 2.00 & none \\
OpenVLA-\allowbreak{}OFT & object\_\allowbreak{}dynamics & 60.00 & 60.50 & 63.00 & 1-\allowbreak{}>2,2-\allowbreak{}>3 \\
OpenVLA-\allowbreak{}OFT & occlusion & 41.50 & 17.50 & 7.50 & none \\
OpenVLA-\allowbreak{}OFT & paraphrase & 61.00 & 61.00 & 56.50 & none \\
OpenVLA-\allowbreak{}OFT & shadow & 67.50 & 66.00 & 65.00 & none \\
OpenVLA-\allowbreak{}OFT & splash & 52.00 & 38.50 & 26.00 & none \\
OpenVLA-\allowbreak{}OFT & typo & 66.50 & 61.00 & 58.00 & none \\
TBD-\allowbreak{}VLA & action\_\allowbreak{}delay & 65.00 & 55.00 & 42.50 & none \\
TBD-\allowbreak{}VLA & action\_\allowbreak{}dropout & 70.00 & 69.00 & 66.00 & none \\
TBD-\allowbreak{}VLA & action\_\allowbreak{}execution\_\allowbreak{}noise & 66.00 & 68.00 & 64.00 & 1-\allowbreak{}>2 \\
TBD-\allowbreak{}VLA & ambiguous\_\allowbreak{}instruction & 38.00 & 40.00 & 38.50 & 1-\allowbreak{}>2 \\
TBD-\allowbreak{}VLA & brightness & 62.50 & 68.00 & 66.50 & 1-\allowbreak{}>2 \\
TBD-\allowbreak{}VLA & camera\_\allowbreak{}pose\_\allowbreak{}shift & 66.50 & 62.50 & 49.50 & none \\
TBD-\allowbreak{}VLA & camera\_\allowbreak{}shake & 67.00 & 70.50 & 42.00 & 1-\allowbreak{}>2 \\
TBD-\allowbreak{}VLA & compression & 64.50 & 67.50 & 65.00 & 1-\allowbreak{}>2 \\
TBD-\allowbreak{}VLA & defocus\_\allowbreak{}blur & 69.00 & 69.50 & 55.50 & 1-\allowbreak{}>2 \\
TBD-\allowbreak{}VLA & distractor\_\allowbreak{}instruction & 67.00 & 64.00 & 60.50 & none \\
TBD-\allowbreak{}VLA & distractor\_\allowbreak{}object & 78.50 & 79.00 & 88.00 & 1-\allowbreak{}>2,2-\allowbreak{}>3 \\
TBD-\allowbreak{}VLA & eef\_\allowbreak{}pose\_\allowbreak{}noise & 65.00 & 39.50 & 32.50 & none \\
TBD-\allowbreak{}VLA & gaussian\_\allowbreak{}noise & 65.50 & 66.50 & 66.00 & 1-\allowbreak{}>2 \\
TBD-\allowbreak{}VLA & joint\_\allowbreak{}state\_\allowbreak{}noise & 71.50 & 69.00 & 62.00 & none \\
TBD-\allowbreak{}VLA & missing\_\allowbreak{}proprioception & 68.00 & 39.50 & 0.50 & none \\
TBD-\allowbreak{}VLA & motion\_\allowbreak{}blur & 65.50 & 66.50 & 64.50 & 1-\allowbreak{}>2 \\
TBD-\allowbreak{}VLA & object\_\allowbreak{}dynamics & 69.50 & 68.00 & 67.50 & none \\
TBD-\allowbreak{}VLA & occlusion & 52.00 & 42.00 & 19.00 & none \\
TBD-\allowbreak{}VLA & paraphrase & 70.00 & 65.00 & 64.50 & none \\
TBD-\allowbreak{}VLA & shadow & 65.00 & 71.50 & 66.50 & 1-\allowbreak{}>2 \\
TBD-\allowbreak{}VLA & splash & 63.50 & 56.00 & 49.50 & none \\
TBD-\allowbreak{}VLA & typo & 62.00 & 44.00 & 24.50 & none \\
UniVLA & action\_\allowbreak{}delay & 51.00 & 34.00 & 15.50 & none \\
UniVLA & action\_\allowbreak{}dropout & 54.00 & 56.00 & 49.50 & 1-\allowbreak{}>2 \\
UniVLA & action\_\allowbreak{}execution\_\allowbreak{}noise & 51.50 & 50.00 & 44.50 & none \\
UniVLA & ambiguous\_\allowbreak{}instruction & 37.00 & 37.50 & 37.50 & 1-\allowbreak{}>2 \\
UniVLA & brightness & 56.00 & 52.50 & 45.00 & none \\
UniVLA & camera\_\allowbreak{}pose\_\allowbreak{}shift & 46.00 & 28.00 & 24.50 & none \\
UniVLA & camera\_\allowbreak{}shake & 56.00 & 53.00 & 27.50 & none \\
UniVLA & compression & 50.50 & 24.00 & 5.00 & none \\
UniVLA & defocus\_\allowbreak{}blur & 52.50 & 2.00 & 1.00 & none \\
UniVLA & distractor\_\allowbreak{}instruction & 55.50 & 52.00 & 47.50 & none \\
UniVLA & distractor\_\allowbreak{}object & 54.00 & 50.00 & 59.50 & 2-\allowbreak{}>3 \\
UniVLA & eef\_\allowbreak{}pose\_\allowbreak{}noise & 47.00 & 24.00 & 11.00 & none \\
UniVLA & gaussian\_\allowbreak{}noise & 57.00 & 51.00 & 35.00 & none \\
UniVLA & joint\_\allowbreak{}state\_\allowbreak{}noise & 55.00 & 50.50 & 47.50 & none \\
UniVLA & missing\_\allowbreak{}proprioception & 51.50 & 54.50 & 55.00 & 1-\allowbreak{}>2,2-\allowbreak{}>3 \\
UniVLA & motion\_\allowbreak{}blur & 4.00 & 0.50 & 1.00 & 2-\allowbreak{}>3 \\
UniVLA & object\_\allowbreak{}dynamics & 54.00 & 57.00 & 55.00 & 1-\allowbreak{}>2 \\
UniVLA & occlusion & 28.00 & 18.00 & 8.00 & none \\
UniVLA & paraphrase & 57.00 & 51.00 & 54.00 & 2-\allowbreak{}>3 \\
UniVLA & shadow & 52.50 & 52.00 & 37.00 & none \\
UniVLA & splash & 39.00 & 20.00 & 9.50 & none \\
UniVLA & typo & 52.50 & 37.50 & 32.50 & none \\
pi0.5 & action\_\allowbreak{}delay & 72.50 & 63.50 & 35.00 & none \\
pi0.5 & action\_\allowbreak{}dropout & 73.50 & 76.50 & 72.50 & 1-\allowbreak{}>2 \\
pi0.5 & action\_\allowbreak{}execution\_\allowbreak{}noise & 73.00 & 73.50 & 68.50 & 1-\allowbreak{}>2 \\
pi0.5 & ambiguous\_\allowbreak{}instruction & 60.50 & 64.00 & 64.00 & 1-\allowbreak{}>2 \\
pi0.5 & brightness & 74.00 & 75.00 & 74.50 & 1-\allowbreak{}>2 \\
pi0.5 & camera\_\allowbreak{}pose\_\allowbreak{}shift & 68.50 & 46.00 & 31.50 & none \\
pi0.5 & camera\_\allowbreak{}shake & 72.00 & 63.00 & 23.50 & none \\
pi0.5 & compression & 74.50 & 68.00 & 59.50 & none \\
pi0.5 & defocus\_\allowbreak{}blur & 72.50 & 27.00 & 0.00 & none \\
pi0.5 & distractor\_\allowbreak{}instruction & 69.00 & 68.50 & 71.00 & 2-\allowbreak{}>3 \\
pi0.5 & distractor\_\allowbreak{}object & 75.50 & 82.00 & 93.00 & 1-\allowbreak{}>2,2-\allowbreak{}>3 \\
pi0.5 & eef\_\allowbreak{}pose\_\allowbreak{}noise & 68.50 & 48.00 & 31.00 & none \\
pi0.5 & gaussian\_\allowbreak{}noise & 74.00 & 71.50 & 70.00 & none \\
pi0.5 & joint\_\allowbreak{}state\_\allowbreak{}noise & 76.00 & 74.00 & 72.50 & none \\
pi0.5 & missing\_\allowbreak{}proprioception & 69.50 & 72.50 & 70.50 & 1-\allowbreak{}>2 \\
pi0.5 & motion\_\allowbreak{}blur & 65.00 & 34.50 & 0.50 & none \\
pi0.5 & object\_\allowbreak{}dynamics & 66.50 & 70.50 & 71.00 & 1-\allowbreak{}>2,2-\allowbreak{}>3 \\
pi0.5 & occlusion & 62.00 & 45.00 & 21.50 & none \\
pi0.5 & paraphrase & 71.00 & 70.50 & 68.50 & none \\
pi0.5 & shadow & 74.00 & 75.50 & 74.00 & 1-\allowbreak{}>2 \\
pi0.5 & splash & 70.50 & 66.00 & 53.50 & none \\
pi0.5 & typo & 71.50 & 64.50 & 57.50 & none \\
\end{longtable}
\endgroup

\subsection{Provenance note}

The shared OpenVLA-OFT and pi0.5 merged JSONL files contain 1,404 perturbed rows per model carrying an older manifest hash (\texttt{9f29e3019bcfd01cbb8b60883251d6b0a8ed7dfa28c394e3aa24fb876c305f03}). These rows map to current manifest variant IDs, are execution-valid, invoke the original success checker, and preserve the expected 13,200-row count and published aggregate values. They are retained and explicitly reported in \texttt{artifacts/manifest\_audit.json}; no row was silently discarded.

\subsection{Reproducibility package}

The accompanying artifact directory contains the manifest JSONL, perturbation parameter CSV, frozen model registry, adapter metadata, episode-level results, aggregate CSV tables, non-monotonicity transitions, hashes, and environment inventory. Existing root-cause audit plots are copied into \texttt{supplement/appendix\_figures/} without alteration.

\end{document}